\documentclass[journal]{IEEEtai}

\usepackage{microtype}
\usepackage{graphicx}
\usepackage{booktabs} 

\usepackage{subcaption} 
\usepackage{multirow}

\let\labelindent\relax
\usepackage{enumitem} 

\usepackage{amsthm}

\usepackage[usenames,dvipsnames,svgnames,table]{xcolor}
\usepackage{xspace}

\usepackage{amsmath,amssymb,amsthm}
\usepackage{amsfonts}

\usepackage{algorithm}
\usepackage{algpseudocode}
\algrenewcommand\algorithmicrequire{\textbf{Input:}}
\algrenewcommand\algorithmicensure{\textbf{Output:}}
\usepackage{tabularx}
\makeatletter
\newcommand{\multiline}[1]{%
  \begin{tabularx}{\dimexpr\linewidth-\ALG@thistlm}[t]{@{}X@{}}
    #1
  \end{tabularx}
}
\makeatother

\usepackage[pagebackref=true,breaklinks=true,letterpaper=true,colorlinks,bookmarks=false]{hyperref}

\def\onedot{\ifx\@let@token.\else.\null\fi\xspace}
\def\eg{\emph{e.g}\onedot} 
\def\ie{\emph{i.e}\onedot} 
\def\wrt{w.r.t\onedot} 
\def\etal{\emph{et al}\onedot}

\def\Vec#1{{\boldsymbol{#1}}} 
\def\Mat#1{{\boldsymbol{#1}}} 
 
\def\GRASS#1#2{\mathcal{G}({#2},{#1})} 
\def\ST#1#2{\mathrm{St}({#2},{#1})} 
\def\ORTHO#1{\mathcal{O}({#1})} 

\begin{document}

\title{Unsupervised Deep Metric Learning via Orthogonality based Probabilistic Loss}

\author{Ujjal Kr Dutta, 
        Mehrtash Harandi, 
        and Chellu Chandra Sekhar
        

\thanks{Recommended for acceptance in the IEEE Transactions on Artificial Intelligence (IEEE TAI).}
}


\maketitle

\begin{abstract}
Metric learning is an important problem in machine learning. It aims to group similar examples together. Existing state-of-the-art metric learning approaches require class labels to learn a metric. As obtaining class labels in all applications is not feasible, we propose an unsupervised approach that learns a metric without making use of class labels. The lack of class labels is compensated by obtaining pseudo-labels of data using a graph-based clustering approach. The pseudo-labels are used to form triplets of examples, which guide the metric learning. We propose a probabilistic loss that minimizes the chances of each triplet violating an angular constraint. A weight function, and an orthogonality constraint in the objective speeds up the convergence and avoids a model collapse. We also provide a stochastic formulation of our method to scale up to large-scale datasets. Our studies demonstrate the competitiveness of our approach against state-of-the-art methods. We also thoroughly study the effect of the different components of our method.
\end{abstract}




\section{Introduction}
\label{intro}

\IEEEPARstart{A} key step in artificial intelligence and machine learning algorithms is to find the distance or similarity among examples. Distance Metric Learning (DML) is equivalent to obtaining an embedding where similar examples are grouped together, while moving away dissimilar ones. Recent works in machine learning problems like few-shot object detection \cite{RepMet_CVPR19}, zero-shot recognition \cite{Adap_DML_ZSL_SPL19}, zero-shot image retrieval and clustering \cite{ECAML_AAAI19}, co-saliency detection \cite{co_saliency_DML_17}, remote sensing \cite{remote_sensing_DML_18} and fine-grained categorization \cite{fgvc_pami19} have demonstrated the benefits of learning a distance metric.

Despite their exemplary performance, the problem with existing state-of-the-art DML approaches is that they are \textit{supervised} in nature, \ie, they require class labels (manual annotations) to learn the metric, often in large scale. However, many applications do not have the feasibility of obtaining supervisory signals or class labels. Examples of such applications include medical imaging techniques requiring invasive procedures \cite{perone2019unsupervised,yang2008boosting}, large-scale 3D point cloud image recognition \cite{Landrieu_2019_CVPR,zhao2020jsnet}, and image segmentation requiring pixel-level annotations \cite{qian2019weakly}, to name a few. Furthermore, obtaining manual annotations also require certain degree of domain expertise, and subjective biases, which often leads to noisy or erroneous labels. Thus, it is essential to be able to learn a metric that could capture the inherent properties of data, without requiring class labels.

To learn a metric, existing supervised DML approaches provide constraints in the form of pairs \cite{contrastive}, triplets \cite{FaceNet}, tuples \cite{N_pair} or batches \cite{Lifted_structure}. These constraints are obtained using the available class labels. Hence, the key challenge that naturally arises in \textit{unsupervised metric learning} is that of obtaining such constraints. Classically, it has been studied in the context of \textit{manifold learning} \cite{LPP,NPE}.
\begin{figure}[t]
\centering
	\includegraphics[width=0.85\columnwidth]{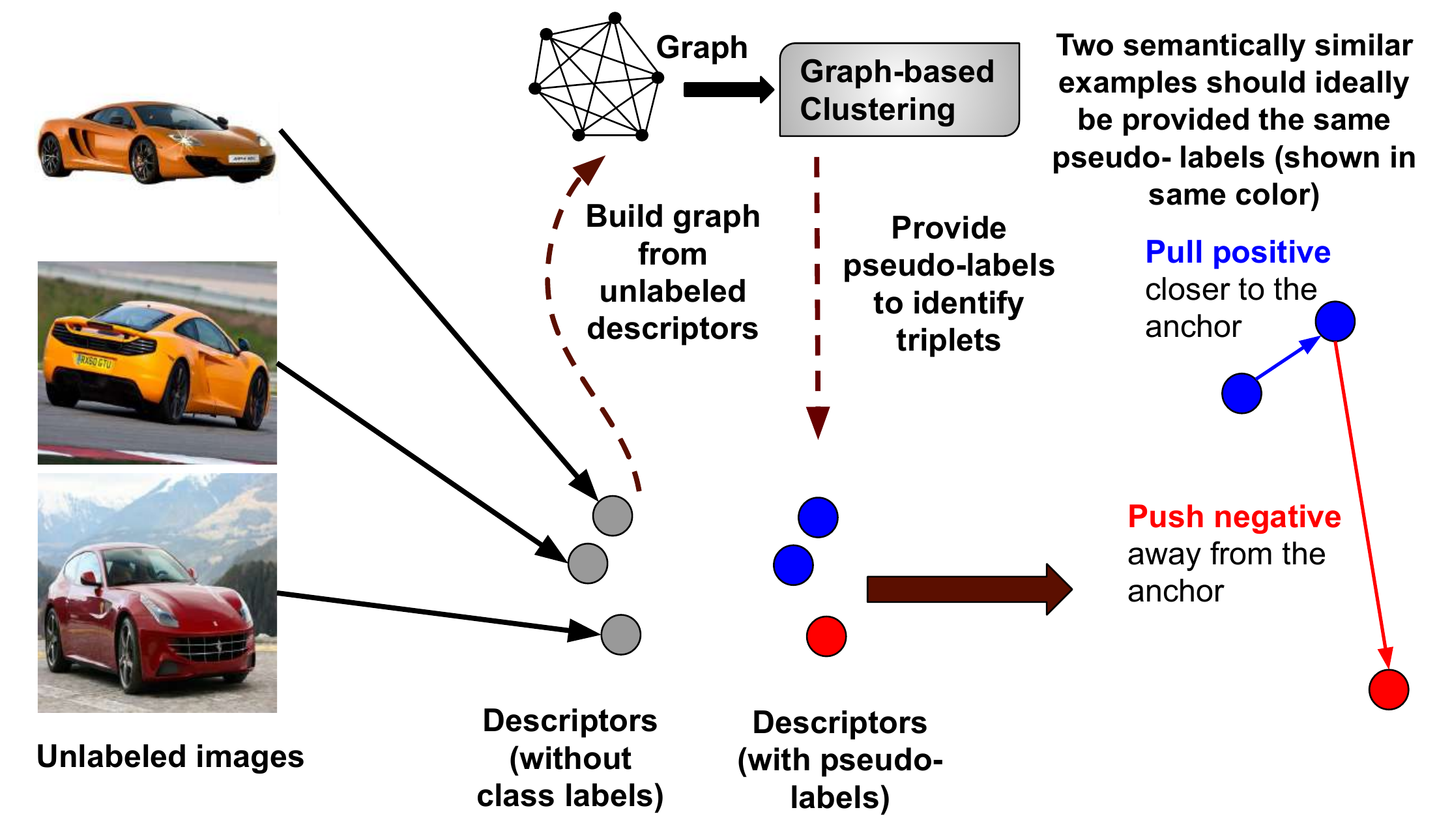}
    \caption{An illustration of the proposed approach. The input images belong to Cars196 dataset \cite{Cars196}. The figure is best viewed in color.}
    \label{framework_OPML}
\end{figure}

A few recent approaches have attempted this problem from a feature learning perspective \cite{DeepCluster_ECCV18,NCE_CVPR18,ExemplarCNN_TPAMI16,InvariantSpread_CVPR19}. Under the absence of any supervision, some of the approaches \cite{NCE_CVPR18,InvariantSpread_CVPR19} have treated each example and its augmentations as a distinct class. 
The Mining on Manifolds (MOM) approach by Iscen \etal \cite{MOM} follows a paradigm for unsupervised metric learning by making use of a graph. In contrast to other feature learning approaches, rather than treating each instance as a separate class, they actually identify groups of examples that are similar or dissimilar to each other. By performing a random-walk on a graph, they identify manifold and Euclidean neighbors to form triplet constraints, thus indicating a notion of similarity among examples.

The intuitive nature of the MOM approach motivates us to leverage graph-based constraint mining. In contrast to MOM that directly mines hard constraints using manifold and Euclidean neighbors, we simply partition a dataset and obtain pseudo-labels for each example, by using a graph-based clustering approach called Authority Ascent Shift (AAS) \cite{AAS}. This is because pseudo labels provide us the flexibility to form any type of constraints as desired. With pseudo label information for a mini-batch, one can directly apply existing supervised techniques in a straightforward manner. Moreover, we can also apply online constraint mining strategies (\eg, semi-hard strategy \cite{FaceNet}), and recent batch based strategies \cite{XBM_DML_CVPR20}. Our approach is illustrated in Figure \ref{framework_OPML}. The pseudo-labels are used to identify triplet constraints for metric learning. A triplet essentially consists of an \textit{anchor-positive} pair of semantically similar examples, and a third \textit{negative} example that is dissimilar to the pair.

Following are the major \textbf{contributions of our paper}: 1. We propose a novel probabilistic metric learning objective to ensure that the triplets satisfy an angular property \cite{Angular_loss}. 2. We further include a weight function, and an orthogonality constraint in our objective. This speeds up the convergence, and avoids a model collapse. 3. We also propose a stochastic formulation of our method to scale up to large datasets, while handling non-linearity in data using deep neural networks. We empirically establish the competitiveness of our method in comparison to existing state-of-the-art approaches, and also carefully analyze the role of different components of our method.

\section{Related Work}
\label{sec_rel_work}

Classically, unsupervised DML has been studied as a by-product of \textit{manifold learning} \cite{LPP,NPE} or \textit{diffusion processes} \cite{donoser2013diffusion,iscen2017efficient}. In the absence of class labels, a natural way to identify the semantic similarity among two examples is to first perform a clustering of the data to obtain \textit{pseudo-labels}, and then learn a metric. The DeepCluster \cite{DeepCluster_ECCV18} approach follows this approach of jointly learning pseudo-labels obtained from traditional k-means clustering, and using these labels to learn an embedding. However, center-based clustering techniques like k-means are well-known for their drawbacks. For example, sensitivity to initialization, setting the number of clusters apriori, and the lack of effectiveness in higher dimensions.

The Exemplar \cite{ExemplarCNN_TPAMI16} method randomly samples a set of image patches. It assumes that a few random parameter vectors represent certain elementary transformation operations like translation, scaling, rotation, contrast, and colorization. The goal is to learn these parameters by following a standard supervised setting, where the classes will be obtained from the transformed patches, with the identity of an example denoting the class label.

The NCE \cite{NCE_CVPR18} and InvariantSpread \cite{InvariantSpread_CVPR19} approaches obtain augmentations of examples, and seek to learn an embedding that pulls augmentations of an example together, while moving away augmentations of different examples. The InvariantSpread method is formulated as an extension of the NCE approach. The Exemplar, NCE, and InvariantSpread methods are \textit{instance-wise} in nature, \ie, they treat each example as a separate class. Ideally, not only we demand augmentations of the same example to be close, we also want to group together two \textit{different} examples, but of the same \textit{semantic similarity}. To address this, the Mining on Manifolds (MOM) \cite{MOM} approach makes use of a graph-based ranking technique by performing a random-walk. By virtue of the underlying ranking, it identifies manifold and Euclidean neighbors to form triplets. In particular, it identifies \textit{hard} positives and negatives with respect to an anchor example. It then learns a metric by using a standard triplet loss.

\section{Background}

Let $\Vec{x}_i \in \mathbb{R}^d$ be the descriptor of an example $i$ in a dataset $\mathcal{X}$, which is unlabeled. For $\Vec{x}_i \in \mathbb{R}^d$, let $\Mat{L}^\top\Vec{x}_i \in \mathbb{R}^l$, denote its learned embedding. Here, $\Mat{L}\in \mathbb{R}^{d \times l}$ is the parametric matrix of the squared Mahalanobis-like distance metric $\delta^2_{\Mat{L}}(\Vec{x}_i,\Vec{x}_j)=(\Vec{x}_i-\Vec{x}_j)^\top\Mat{L}\Mat{L}^\top(\Vec{x}_i-\Vec{x}_j)$, for a pair of examples $\Vec{x}_i,\Vec{x}_j \in \mathbb{R}^d$. Ensuring $l < d$ facilitates dimensionality reduction. The goal of our work is to learn the parametric matrix $\Mat{L}$. As $\mathcal{X}$ is unlabeled, we do not have class labels for learning $\Mat{L}$. To compensate for the lack of class labels, we suggest obtaining pseudo-labels. In our work, we choose the graph-based Authority Ascent Shift (AAS) clustering \cite{AAS} to obtain the pseudo-labels.

Let, the AAS clustering be denoted by a function $c:\mathbb{R}^d\rightarrow\mathbb{Z}^+$ such that $c(\Vec{x}_i)$, a positive integer, denotes the \textit{pseudo-label} assigned to $\Vec{x}_i\in \mathbb{R}^d$. Briefly, AAS requires constructing a weighted graph with nodes representing the examples, and edges between the nearest neighbors. Edge weights denote \textit{affinities} between examples. With $\omega$ denoting the stationary probability distribution of a random walker on the graph, the \textit{node relevancy} from node $i$ to node $j$ can be defined as \cite{AAS}:
\begin{equation}
   \label{node_relevancy}
    \psi(i,j)=d_i\Mat{T}_{ij}\exp(-\gamma(\nabla_{\omega}(i,j))^2).
\end{equation}
Here, $d_i$ is the out-degree of node $i$, $\Mat{T}_{ij}$ is the transition probability from node $i$ to node $j$, $\exp(.)$ is the exponential function, $\nabla_{\omega}(i,j)=[\omega(j)-\omega(i)]$ and $\gamma>0$ is a hyperparameter. The set of \textit{relevant neighbors} of node $i$ can be defined as:
\begin{equation}
   \label{relevant_neighbors}
    \mathcal{N}_\epsilon(i)=\left \{ j\in\mathcal{V}:\psi(i,j)> \epsilon \right \}\cup \left \{ i \right \}.
\end{equation}
Here, $\epsilon>0$ is a hyperparameter and $\mathcal{V}$ is the vertex set of the graph. \textit{Authority ascent} of a node $i$ can be performed by moving towards the node $j^*$ such that $j^*=\operatorname*{arg\,max}_{j \in \mathcal{N}_\epsilon(i)} \Mat{T}_{ij}\nabla_{\omega}(i,j)$. By subsequently performing authority ascent on neighboring nodes, we can associate a \textit{authority mode} \cite{AAS} to node $i$. Nodes sharing a common authority mode build a tree. Disjoint trees represent the distinct, arbitrary-shaped clusters present in the data \cite{AAS}.

\textbf{Motivation to use AAS:} AAS does not require to fix the number of clusters apriori, which is a crucial benefit in the unsupervised setting. It is able to detect arbitrary-shaped clusters in the data, while being robust to noise and outliers. Li \etal \cite{Cyclic_ECCV16} pointed the necessity to capture \textit{intra-class variances} that may occur in visual data due to minor pose, illumination or viewpoint differences. Such variances can be captured by AAS, as it relies on a notion of geometric similarity.

\section{Proposed Methodology}

Using the clustering function $c(.)$, we can form a set of triplets: $\mathcal{T} =\{ (\Vec{x}_i,\Vec{x}_i^+,\Vec{x}_i^-) \}_{i=1}^{|\mathcal{T}|}$, each element of which consists of the following: i) $\Vec{x}_i$, an arbitrary example with a value $c(\Vec{x}_i)$
. ii) $\Vec{x}_i^+$, another arbitrary example with $c(\Vec{x}_i^+)=c(\Vec{x}_i)$. iii) $\Vec{x}_i^-$, such that $c(\Vec{x}_i^-) \neq c(\Vec{x}_i)$. The examples $\Vec{x}_i$, $\Vec{x}_i^+$ and $\Vec{x}_i^-$ are referred to as the \textit{anchor}, \textit{positive} and \textit{negative} respectively (Figure \ref{framework_OPML}). Here, $|\mathcal{T}|$ is the number of triplets formed. Using $\mathcal{T}$, our goal is to learn $\Mat{L}$ in $\delta^2_{\Mat{L}}(\Vec{x}_i,\Vec{x}_j)$. In particular, we make use of the \textit{semi-hard} strategy \cite{FaceNet} of mining triplets using the pseudo-labels.

\subsection{Probabilistic Metric Learning Objective with Orthogonality}
Given a triplet $(\Vec{x}_i,\Vec{x}_i^+,\Vec{x}_i^-)$, we seek to minimize the following \textit{angular constraint} \cite{Angular_loss} based hinge-loss term:
\begin{equation}
\label{eqn_z_i}
[z_i]_+=[\delta^2_{\Mat{L}}(\Vec{x}_i,\Vec{x}_i^+)-4\textrm{ tan}^2\alpha \textrm{ } \delta^2_{\Mat{L}}(\Vec{x}_i^-,\Vec{x}_{i-avg})]_+.
\end{equation}

\begin{figure}[!htb]
\centering
\includegraphics[width=0.8\columnwidth]{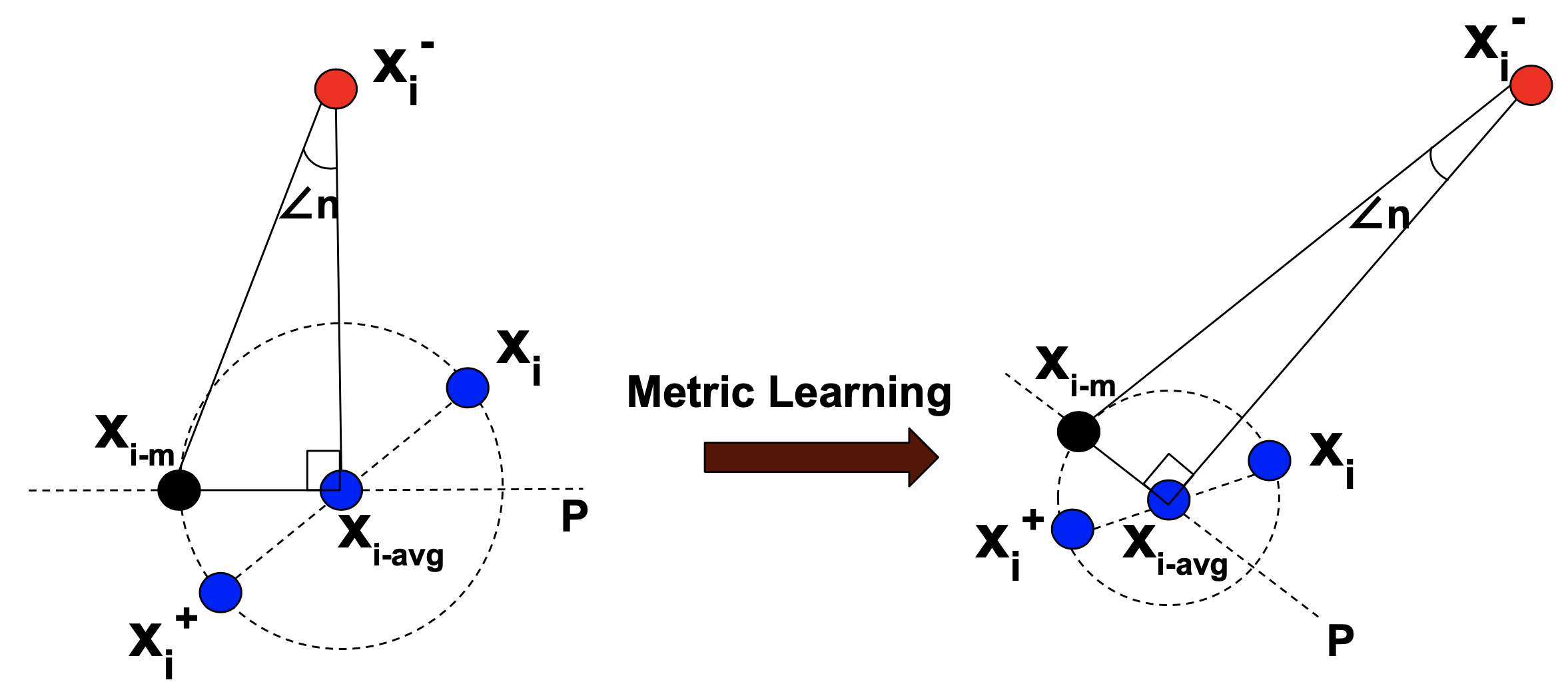}
\caption{Illustration of the angular constraint.}
\label{angularloss_illus}
\end{figure}
Here, $[z_i]_+=\textrm{max}(0,z_i)$. Figure \ref{angularloss_illus} illustrates the angular constraint using a circle centered at $\Vec{x}_{i-avg}=\frac{\Vec{x}_i+\Vec{x}_i^+}{2}$, while having the points $\Vec{x}_i$ and $\Vec{x}_i^+$ on its perimeter, and $\Vec{x}_i^-$ outside it. The line $P$ passing through $\Vec{x}_{i-avg}$ intersects the circle at $\Vec{x}_{i-m}$, and is perpendicular to the line segment $\Vec{x}_{i-avg}-\Vec{x}_i^-$. By constraining the $\angle n$ with an upper bound $\angle \alpha >0^\circ$ (a hyperparameter), one can push the negative $\Vec{x}_i^-$ away from the center $\Vec{x}_{i-avg}$ of the local cluster defined by $\Vec{x}_i$ and $\Vec{x}_i^+$, while dragging the latter two closer. This is achieved by constraining: $\textrm{ tan} (n) \leq \textrm{ tan} (\alpha) \Rightarrow \delta_{\Mat{L}}(\Vec{x}_{i-m},\Vec{x}_{i-avg})/ \delta_{\Mat{L}}(\Vec{x}_i^-,\Vec{x}_{i-avg}) \leq \textrm{ tan} (\alpha) \Rightarrow 0.5\delta_{\Mat{L}}(\Vec{x}_i,\Vec{x}_i^+)/ \delta_{\Mat{L}}(\Vec{x}_i^-,\Vec{x}_{i-avg}) \leq \textrm{ tan} (\alpha) \Rightarrow
\delta^2_{\Mat{L}}(\Vec{x}_i,\Vec{x}_i^+) \leq 4\textrm{ tan}^2\alpha \textrm{ } \delta^2_{\Mat{L}}(\Vec{x}_i^-,\Vec{x}_{i-avg})$, which leads to the loss in (\ref{eqn_z_i}).

As $[z_i]_+$ is non-smooth, we propose to minimize the following smooth version: $m_i=\log(1+\exp(z_i))$ instead of $[z_i]_+$. This is possible as $\log(\exp(a)+\exp(b))\geq \textrm{max}(a,b)$ where $a,b\in \mathbb{R}$. Let, $\sigma(a)=\frac{1}{1+\exp(-a)}, a\in \mathbb{R}$ be the logistic function. We now formulate a novel probabilistic metric learning objective to satisfy the above angular constraint in $m_i$. For this, we define the probability of a triplet $(\Vec{x}_i,\Vec{x}_i^+,\Vec{x}_i^-)$ not violating the angular constraint, as follows:
\begin{equation}
\label{p_i}
\begin{split}
    &p\textrm{\{}(\Vec{x}_i,\Vec{x}_i^+,\Vec{x}_i^-) \textrm{ does not violate the angular constraint\}}\\
    &=p_i=\sigma(-f_i).
\end{split}
\end{equation}
Here, $f_i$ is a \textit{weighted loss} defined as: $f_i=w_i m_i$. Note that minimizing the loss $f_i$ will maximize $p_i$. $w_i$ is a weight term on the smooth loss $m_i$, which is defined as a function $w_i:\mathbb{R}^d \times \mathbb{R}^d \times \mathbb{R}^d \rightarrow [0,1]$ that provides a scaling weight for the loss term on a triplet. $w_i$ has a parametric matrix $\Mat{R}\in\mathbb{R}^{d\times l}$, which we discuss later.

To ensure that a triplet satisfies the angular constraint, we need to maximize $p_i$. As each triplet is independent, the joint probability that all the triplets satisfy the angular constraint is expressed as $\prod_ip_i$. Thus, the log likelihood of the triplet set $\mathcal{T}$ satisfying the angular constraint is expressed as: $\log\prod_ip_i=\sum_i\log p_i$. To learn the distance metric, we propose to maximize this log likelihood (\ie, \textit{minimizing} the negative log likelihood), by solving the following optimization problem:
\begin{equation}
\label{OPML}
\min_{\Mat{R},\Mat{L}} \mathcal{L}(\Mat{R},\Mat{L})=\sum_{i=1}^{|\mathcal{T}|}(-\log p_i).
\end{equation}

Let, the learned embeddings of the anchor and the positive be denoted as $\Mat{L}^\top\Vec{x}_i$ and $\Mat{L}^\top\Vec{x}_i^+$ respectively. The bilinear similarity between them can be expressed as $(\Mat{L}^\top\Vec{x}_i)^\top \Mat{L}^\top\Vec{x}_i^+=\Vec{x}_i^\top\Mat{L}\Mat{L}^\top\Vec{x}_i^+$. A lower value of similarity indicates that the pair is too hard to be an anchor-positive one, and should not have been grouped together by the clustering. Hence, we should down-weigh the loss associated with this triplet via a weight term $w_i^+$.

On the contrary, $\Vec{x}_{i-avg}^\top\Mat{L}\Mat{L}^\top\Vec{x}_i^-$ represents the bilinear similarity of the negative $\Vec{x}_i^-$, \wrt the average representation $\Vec{x}_{i-avg}$ of the (anchor, positive) pair in the triplet. A higher similarity indicates that all three of them should have been grouped together by the clustering. Hence, we down-weigh the loss associated with this triplet using a weight term $w_i^-$.

We express the final weight term $w_i$ for defining $f_i$ in (\ref{p_i}), as follows: $w_i=(w_i^++w_i^-)/2$. However, parameterizing the function $w_i$ in terms of $\Mat{L}$ is too restrictive on $\Mat{L}$. Hence, we use a separate parametric matrix $\Mat{R}$ of the same dimensions, to represent $\Mat{L}$. We finally encode the bilinear similarities using the logistic function, and define $w_i^+$ and $w_i^-$ as follows: $w_i^+=\sigma(\Vec{x}_i^\top\Mat{R}\Mat{R}^\top\Vec{x}_i^+)$, and $w_i^-=1-\sigma(\Vec{x}_{i-avg}^\top\Mat{R}\Mat{R}^\top\Vec{x}_i^-)$. In the expression for $w_i^-$, the subtraction from $1$ has been done because a higher value of $\Vec{x}_{i-avg}^\top\Mat{R}\Mat{R}^\top\Vec{x}_i^-$ indicates that we should give a lower weightage.
\begin{figure}[t]
\centering
	\includegraphics[width=0.6\columnwidth]{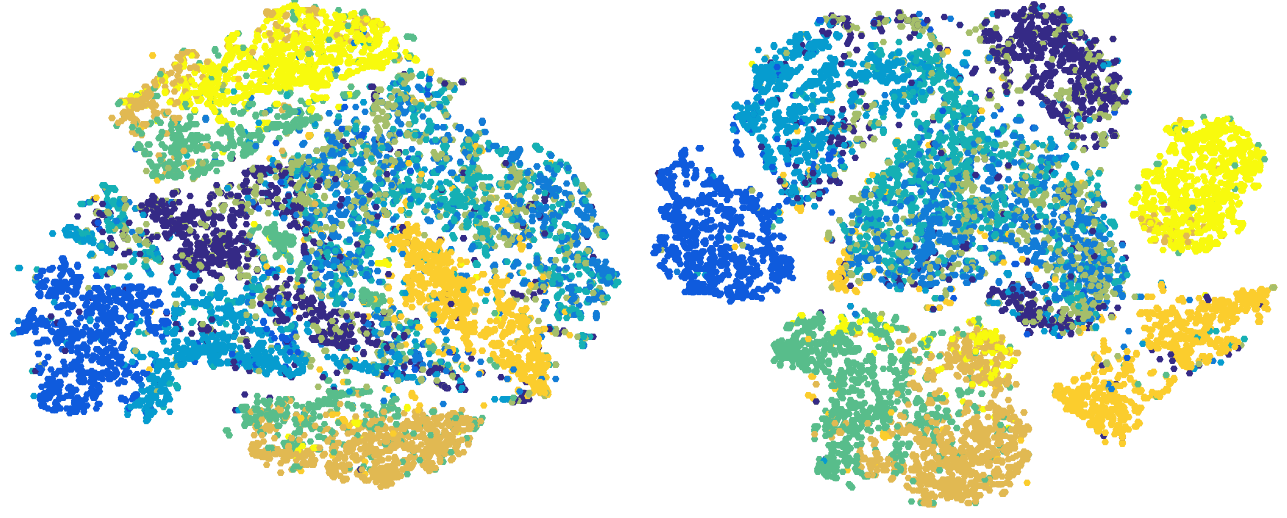}
    \caption{Embedding learned on the standard test split of the Fashion-MNIST dataset \cite{Fashion-MNIST} without orthogonality (left-panel), and with orthogonality (right panel) on the metric parameters. While there may be a model collapse without orthogonality, imposing orthogonality leads to well-separated and compact groups.}
    \label{ortho_motiv}
\end{figure}

A common practice in metric learning is to impose a regularizer on the parameters of the metric. In our work, we impose orthogonality on the matrix $\Mat{L}$. This is because without orthogonality, there may be a model collapse, leading to degenerate embeddings. Figure \ref{ortho_motiv} shows that the embeddings obtained for the standard test split of the Fashion-MNIST dataset \cite{Fashion-MNIST} are relatively better with orthogonality. In the context of metric learning, Xie \etal \cite{MDMLCLDD} further pointed that orthogonality helps in learning a compact set of projection vectors, helps reducing the adverse affects of class imbalance, and avoids overfitting. The general benefits of orthogonality has also been studied by many recent works \cite{soumava_cvpr18,bansal2018can,MDMLCLDD}.
\begin{algorithm}[H]
\begin{scriptsize}
\caption{stochastic OPML (sOPML)}
\label{alg_sOPML}
\begin{algorithmic}[1]
\State {\bfseries Input:} Unlabeled data $\mathcal{X}$, initial $\Phi$; $c:\mathbb{R}^d\rightarrow\mathbb{Z}^+$; $\alpha$, $maxiter>0$.
\State Perform $c(.)$ on $\mathcal{X}$ to obtain pseudo-labels.
\State Initialize $\Mat{R}_{\textrm{prev}}, \Mat{L}_{\textrm{prev}}$.
\While {not converged}
\For{MB {\bfseries in} $data loader$} \Comment{MB: Mini-Batch}
\State images, pseudo-labels $\gets$ MB. \Comment{$|MB|$: batch size}
\State $\{ \Vec{x}_j, c(\Vec{x}_j) \}_{j=1}^{|MB|} \gets (\Phi$(images), pseudo-labels).
\State $\mathcal{T} =\{ (\Vec{x}_i,\Vec{x}_i^+,\Vec{x}_i^-) \}_{i=1}^{|\mathcal{T}|} \gets$  Semi-Hard-Mine($\{ \Vec{x}_j, c(\Vec{x}_j) \}_{j=1}^{|MB|})$ \cite{FaceNet}.
\State $[\Mat{R}, \Mat{L}] \gets$ RCGD($maxiter, \mathcal{T}, \mathcal{L}(\Mat{R}_{\textrm{prev}},\Mat{L}_{\textrm{prev}})$) \Comment{Using (\ref{OPML})}
\State $\Mat{R}_{\textrm{prev}} \gets \Mat{R}$; $\Mat{L}_{\textrm{prev}} \gets \Mat{L}$.
\State loss $\gets \mathcal{L}(\Mat{R},\Mat{L}).$  \Comment{Using (\ref{OPML})}
\State loss.backward() and optimizer.step() to learn $\Phi$. \Comment{BackPropagation}
\EndFor\label{for_outer}
\EndWhile
\State \textbf{return} $\Mat{L}$
\end{algorithmic}
\end{scriptsize}
\end{algorithm}

Upon adding the \textit{orthogonality constraint} $\Mat{L}^\top\Mat{L}=\mathbf{I}_l\in\mathbb{R}^{l\times l}$, the matrix $\Mat{L}$ lies on a orthogonal Stiefel manifold $\ST{d}{l}$ \cite{AMS09}. However, the objective of (\ref{OPML}) is invariant to the right action of the orthogonal group $\ORTHO{l}=\{\Mat{B}\in \mathbb{R}^{l\times l}:\Mat{B}\Mat{B}^\top=\Mat{B}^\top\Mat{B}=\mathbf{I}_l\}$, \ie, for $\Mat{B}\in \ORTHO{l}$, we have $\mathcal{L}(\Mat{R},\Mat{L})=\mathcal{L}(\Mat{R},\Mat{L}\Mat{B})$. Therefore, to jointly learn the parameters $\Mat{R}$ and $\Mat{L}$, we must constrain the optimization problem in (\ref{OPML}) on the following Riemannian product manifold: $\mathcal{M}_{p}\triangleq\mathbb{R}^{d\times l}\times \GRASS{d}{l}$. Here, $\GRASS{d}{l}$ is the Grassmann manifold \cite{AMS09}, which is the quotient of $\ST{d}{l}$. A detailed description on Riemannian geometry and related optimization can be found in \cite{AMS09}. We perform Riemannian Conjugate Gradient Descent (RCGD) to jointly learn the parameters of our method. We call our method as \textbf{Orthogonality based Probabilistic Unsupervised Metric Learning (OPML)}.

\subsection{Incorporation with deep neural networks for scaling up, and computational complexity}
\begin{figure}[!htb]
\centering
	\includegraphics[width=0.8\linewidth]{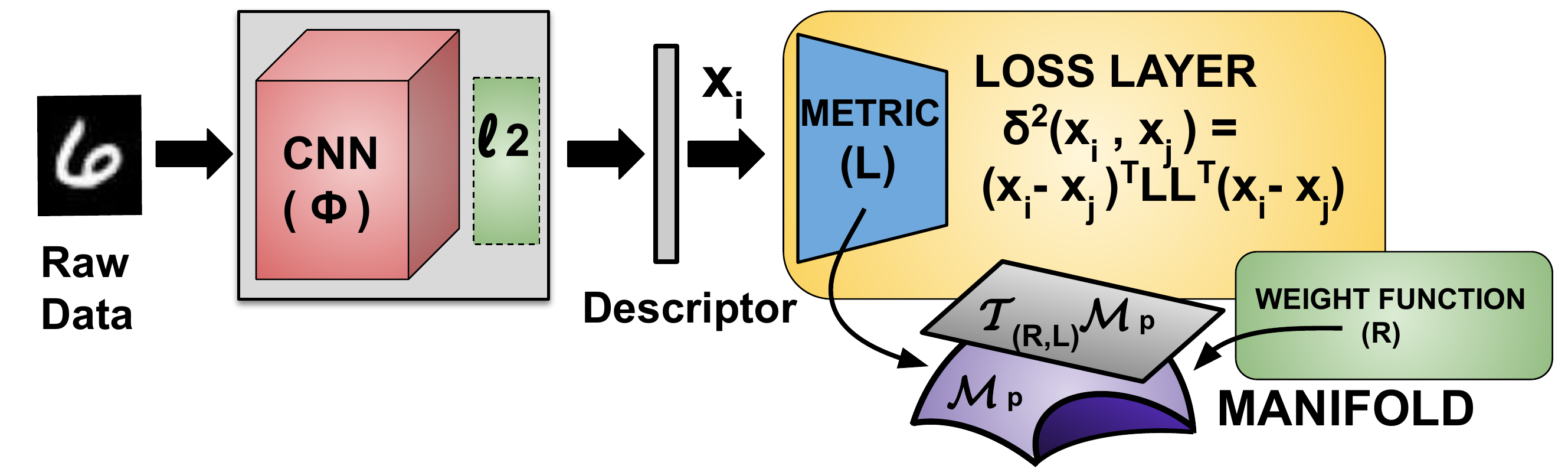}
    \caption{Architecture required for stochastic OPML (sOPML). The raw image belongs to the MNIST \cite{MNIST} dataset. }
    \label{OPML_architecture}
\end{figure}

To handle large datasets, while capturing non-linearity using a Convolutional Neural Network (CNN), we suggest a stochastic formulation of OPML (Algorithm \ref{alg_sOPML}). Given the unlabeled dataset $\mathcal{X}$, let $\Phi$ denote the parameters of a CNN $z:\mathcal{X}\rightarrow \mathbb{R}^d$ that provides the non-linear embedding $\Vec{x}_i \in \mathbb{R}^d$ of the $i^{th}$ raw example in $\mathcal{X}$, for use in (\ref{OPML}). One can perform RCGD to learn $\Mat{L}$ and $\Mat{R}$ within the loss layer, without requiring any modifications in the architecture (see Figure \ref{OPML_architecture}). As our objective is smooth, and fully differentiable, the gradients can be back-propagated to jointly learn $(\Phi,\Mat{R},\Mat{L})$.

The computational complexity of AAS \cite{AAS} depends on the number of power iterations required for computing $\omega$. As pointed by the authors, for real applications with sparse graphs, the complexity is much lower, and grows roughly linearly with the dataset size. \cite{van2014accelerating} showed that it is possible to obtain t-SNE embeddings of datasets with millions of objects in $O(N \textrm{ log}N)$. Using that as a preceding step before performing AAS makes AAS scalable to large datasets, while achieving fairly good clustering results. Alternately, one may randomly sample a sufficiently large partition of the data, and perform AAS on that partition. During training we can first pick a random partition, and sample mini-batches from within the partition. Also, (\ref{OPML}) is linear in $|\mathcal{T}|$ and quadratic in terms of $d$, and hence, efficient as well.

\section{Experimental Results}
In this section, we study the different components of our proposed method, and make empirical comparisons against state-of-the-art approaches on a number of benchmark datasets.

\textbf{Performance Metrics}: All comparisons in our experiments are made with respect to either the clustering or the retrieval performance of an approach on a dataset. The performance on the clustering task is measured in terms of Normalized Mutual Information (NMI), F-measure (F), Precision (P) and Recall (R). NMI is defined as the ratio of mutual information and the average entropy of clusters and entropy of actual ground truth class labels. F-measure is the harmonic mean of precision and recall. For the retrieval task, we use the Recall@K metric that gives us the percentage of test examples that have at least one K nearest neighbor from the same class. A higher value of a metric indicates a better performance for an approach.

\textbf{Datasets}:
The following datasets have been used:
\begin{itemize}
    \item \textbf{MNIST} \cite{MNIST}: It consists of 70000 gray-scale images of handwritten digits, of $28\times 28$ pixels. The standard split consists of 60000 training images and 10000 test images.
    \item \textbf{JHMDB} \cite{JHMDB}: It is an action recognition dataset that consists of 960 video sequences belonging to 21 actions.
    \item \textbf{HMDB} \cite{HMDB}: It is another action recognition dataset that consists of 6766 videos belonging to 51 classes.
    \item \textbf{Animals with Attributes 2 (AwA2)} \cite{ZSL_good_bad_ugly}: It consists of 37322 images belonging to 50 different classes of animals.
    \item \textbf{Fashion-MNIST} \cite{Fashion-MNIST}: It consists of images from 10 categories of fashion products. The standard split is as follows: 60000 training images, and 10000 test images.
    \item \textbf{CUB-200} \cite{CUB}: It consists of images of 200 species of birds with first 100 species for training (5864 examples) and remaining for testing (5924 examples).
    \item \textbf{Cars-196} \cite{Cars196}: This dataset consists of images of cars belonging to 196 models. Usually, the first 98 models containing 8054 images are used for training. The remaining 98 models containing 8131 images are used for testing.
    \item \textbf{Stanford Online Products (SOP)} \cite{Lifted_structure}: This dataset consists of images of online products belonging to 22634 product classes, with a total of 120053 images. Usually, the first 11318 classes containing 59551 images are used for training. The remaining 11316 classes containing 60502 images are used for testing.
\end{itemize}

\begin{figure}[!htb]
\centering
	\includegraphics[width=0.8\columnwidth]{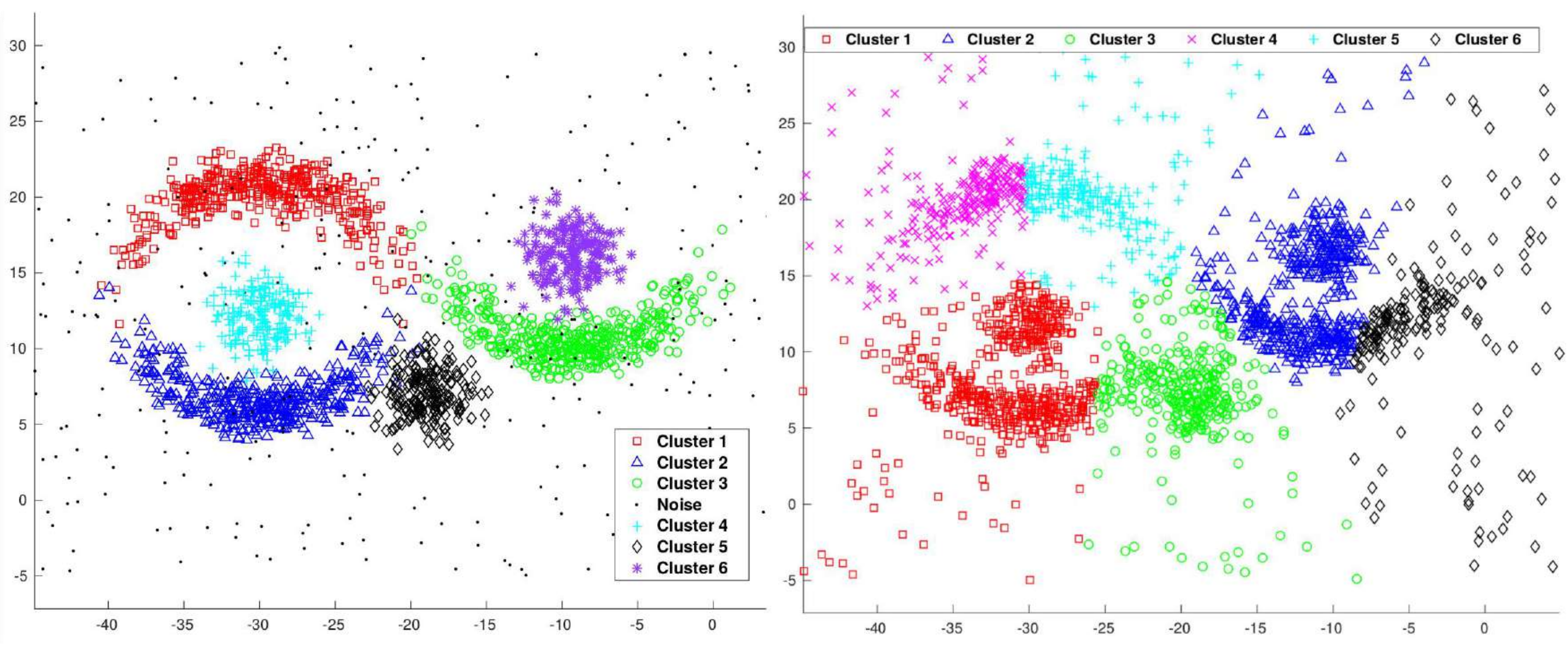}
    \caption{Comparison of AAS (left panel) and k-means (right panel) clustering on a synthetic noisy dataset (Best viewed when zoomed).}
    \label{AAS_vs_kmeans_syn}
\end{figure}

\subsection{Study of the graph-based AAS clustering}
Our proposed approach makes use of the AAS clustering to obtain pseudo-labels for mining triplets. The quality of the learned metric depends on the triplets formed, which in turn are dependent on how well the underlying clustering models the data. Therefore, it is important to study the impact of changing the various hyperparameters of this approach. It is a well known fact that k-means clustering forms spherical clusters due to its center-based model. When data has arbitrary-shaped clusters with uniformly distributed noise, the AAS method is able to capture the data distribution in a better way than k-means.

To validate this claim, we generate a 2D synthetic dataset as shown in Figure \ref{AAS_vs_kmeans_syn}. It consists of 6 classes with 900 examples in total, where 3 classes are of crescent shape, and 3 are Gaussians. There are 300 additional points uniformly distributed to represent noise in the data. As observed in Figure \ref{AAS_vs_kmeans_syn}, k-means clustering does not represent the underlying data distribution. On the other hand, the AAS clustering method models the representation fairly well, in addition to filtering out the noise.

We make use of the available ground truth class labels and obtained cluster labels to empirically evaluate the clustering performance in terms of the NMI and F metrics. The AAS clustering approach obtained an NMI of 80.9 and F-measure of 82.9, whereas, the k-means method obtained NMI and F-measure values of 50.0 and 42.7 respectively.

We also conducted experiments on the standard test split of the benchmark MNIST \cite{MNIST} digit dataset containing 10000 examples. Each image is represented by the concatenation of the raw pixels as a vector, followed by a t-SNE embedding \cite{van2014accelerating}. We observed that the k-means clustering achieves NMI and F-measure of 73.4 and 68.7 respectively, whereas the AAS clustering obtains NMI and F-measure of 77.9 and 71.0 respectively. The performances in the synthetic and MNIST datasets indicate that the AAS method outperforms the k-means clustering approach in certain scenarios.

The AAS clustering approach has the following major hyperparameters: i) $k\_graph$: the number of nearest neighbors in the underlying kNN graph, ii) the scale parameter $\gamma$ in (\ref{node_relevancy}), and iii) the node relevancy threshold $\epsilon$ in (\ref{relevant_neighbors}). The AAS method defines \textit{cluster authority} as the sum of authority scores of all the elements present in a cluster. A cluster with an unusually low cluster authority (\eg, one that contributes less than $\theta_{min}\%$ of the total cluster authority) can be filtered as a \textit{noise cluster}. One may choose to filter out such (less informative/ noisy) clusters by setting an appropriate value of $\theta_{min}$.

For the above experiment in the MNIST dataset, we set the following values for the AAS approach: $\theta_{min}=5$, $k\_graph=500$, $\gamma=10^2$, and $\epsilon=0.9$. For the AAS method, we create a kNN graph with an edge weight defined as: $W_{ij}=\textrm{exp}(-2c^2\frac{\delta^2_{ij}}{\delta^2_{max}})$, where $\delta^2_{ij}$ is the squared Euclidean distance between the examples represented by nodes $i$ and $j$, and $\delta^2_{max}$ is the maximum distance among all the pairs. The scaling constant $c>0$ has been set to 1. For the k-means method, we set the number of clusters equal to the actual number of classes. We now use the same test split of the MNIST dataset and perform further studies to highlight the importance of each of the hyperparameters in AAS.

\begin{table}[!htb]
\centering
\caption{Impact of number of neighbors in the graph used in AAS.}
\label{AAS_kgraph}
\resizebox{0.65\columnwidth}{!}{%
\begin{tabular}{|c|cccc|c|}
\hline
Dataset                  & \multicolumn{5}{c|}{\textbf{MNIST}}                                                                                    \\ \hline \hline
$k\_graph$ & \textbf{NMI}  & \textbf{F}    & \textbf{P}    & \textbf{R}    & \textbf{\begin{tabular}[c]{@{}c@{}}$N\_clusters$\end{tabular}}\\ \hline
50                       & 58.5          & 18.5          & \textbf{75.6} & 10.5          & 214                                                             \\
100                      & 58.1          & 26.4          & 60.5          & 16.9          & 65                                                               \\ \hline
500                      & \textbf{68.0} & \textbf{56.2} & 62.1          & 51.4          & {\underline{13}}                                                          \\ \hline
1000                     & 63.0          & 56.9          & 46.6          & \textbf{73.1} & 7                                                            \\ \hline
\end{tabular}%
}
\end{table}

\subsubsection{Impact of number of neighbors in the graph}
We study the impact of changing the number of nearest neighbors $k\_graph$ in the kNN graph used in AAS. We fix $\gamma=10^2$, $\epsilon=0.5$, and do not eliminate clusters (\ie, $\theta_{min}=0$). The clustering performance of AAS on the MNIST dataset has been shown in Table \ref{AAS_kgraph}, in terms of NMI, F, P and R values. Best performance for each metric is shown in bold against each column. The number of clusters detected is also reported.

We observe that a lower value of $k\_graph$ leads to a better precision, while a higher value of $k\_graph$ leads to a better recall. An intermediate value of $k\_graph$ (around $500$) leads to a balance in the precision and recall metrics. Noticeably, the number of clusters obtained for $k\_graph=500$ is close to the actual number of classes, \ie, $10$. This is indicated by the underline. We also set $k\_graph=5000$, for which all examples get merged to two clusters in MNIST, which is not meaningful.

\subsubsection{Impact of the node relevancy threshold $\epsilon$}
We now fix $k\_graph=500$, $\gamma=10^2$, and study the impact of $\epsilon$ (shown in Table \ref{AAS_epsilon}). Intuitively, a higher value of $\epsilon$ ensures that examples are not readily merged into the same cluster, unless they have sufficient relevance for that cluster. Therefore, we observed that a larger value of $\epsilon$ leads to higher values of performance metrics. However, increasing $\epsilon$ beyond 1 could lead to a sharp rise in the number of clusters (degeneracy because of many examples forming clusters of their own). Therefore, $\epsilon>1$ should be avoided. We observed that for $\epsilon \rightarrow 1$, performing \textit{noise elimination} is a good practice because some of the smaller clusters could be noise or outliers. For the general case of AAS without noise elimination, an intermediate value of $\epsilon$ works well (\eg, $\epsilon=0.65$).
\begin{table}[!htb]
\centering
\caption{Impact of node relevancy threshold in AAS.}
\label{AAS_epsilon}
\resizebox{0.65\columnwidth}{!}{%
\begin{tabular}{|c|cccc|c|}
\hline
Dataset & \multicolumn{5}{c|}{\textbf{MNIST}}\\ \hline \hline
$\epsilon$ & \textbf{NMI}  & \textbf{F}    & \textbf{P}    & \textbf{R}    & \textbf{\begin{tabular}[c]{@{}c@{}}$N\_clusters$\end{tabular}} \\ \hline
0.1     & 66.2          & 55.9          & 62.1          & 50.7          & 13 \\
0.2     & 66.8          & 56.5          & 63.0          & 51.2          & 13 \\
0.5     & 68.0          & 56.2          & 62.1          & 51.4          & 13  \\ \hline
0.6     & 68.6          & 56.5          & 62.2          & \textbf{51.7} & {\underline{13}}  \\ \hline
0.7     & 68.9          & 57.0          & 64.4          & 51.0          & 15 \\
0.8     & 69.6          & 57.1          & 65.4          & 50.7          & 21 \\
0.9     & \textbf{72.7} & \textbf{59.8} & \textbf{81.2} & 47.3          & 53 \\ \hline
\end{tabular}%
}
\end{table}

\subsubsection{Impact of the scale parameter $\gamma$}
We now perform our studies on AAS with respect to the scale parameter $\gamma$. For this, we fix $k\_graph=500$, $\epsilon=0.9$ and also eliminate noise by setting $\theta_{min}=5$. The obtained performances of AAS 
are reported in Table \ref{AAS_gamma}. We observed that for $\gamma=10^2$, the performance is fairly accurate and the number of clusters (shown with underline) is closer to the actual number of classes present in the data. It should be noted that the values of performance metrics reported in Table (\ref{AAS_gamma}) are higher than those reported in Table (\ref{AAS_kgraph}) and Table (\ref{AAS_epsilon}). This is attributed to the noise removal being performed.
\begin{table}[!htb]
\centering
\caption{Impact of scale parameter $\gamma$ on AAS.}
\label{AAS_gamma}
\resizebox{0.55\columnwidth}{!}{%
\begin{tabular}{|c|cccc|c|cccc|c|}
\hline
Dataset                 & \multicolumn{5}{c|}{\textbf{MNIST}}                                                                                           \\ \hline \hline
$\gamma$ & \textbf{NMI}  & \textbf{F}    & \textbf{P}    & \textbf{R}    & \textbf{\begin{tabular}[c]{@{}c@{}}$N\_clusters$\end{tabular}} \\ \hline \hline
$10$                    & 76.7          & 69.4          & 79.8          & 61.3          & 12                                                                \\
$10^2$                  & \textbf{77.9} & 71.0          & \textbf{80.7} & 63.3          & \underline{11}                                                               \\
$10^3$                 & 77.7          & \textbf{77.7} & 65.6          & \textbf{95.3} & 4                                                               \\ \hline
\end{tabular}%
}
\end{table}

\subsubsection{Effect of eliminating noisy clusters}
As part of our last experiments on AAS, we study the impact of eliminating \textit{noisy clusters}. We set $k\_graph=500$, $\epsilon=0.9$ and $\gamma=10$. Now we vary the values of $\theta_{min}$ and report our observations in Table \ref{AAS_noise}. Here, $\theta_{min}=0$ indicates that we are not eliminating any noisy cluster. We observed that for the same values of fixed hyperparameters, simply eliminating noisy clusters leads to a better clustering performance by AAS. Typically, each of the individual clusters contribute to roughly around $10-20\%$ of the total authority score. Thus setting $\theta_{min}>20$ might falsely eliminate all the clusters as noise. Usually, the range $\theta_{min}\in [1,5]$ leads to meaningful clustering results.
\begin{table}[!htb]
\centering
\caption{Impact of noise removal on AAS.}
\label{AAS_noise}
\resizebox{0.58\columnwidth}{!}{%
\begin{tabular}{|c|cccc|c|}
\hline
Dataset                   & \multicolumn{5}{c|}{\textbf{MNIST}}                                                                                               \\ \hline \hline
$\theta_{min}$ & \textbf{NMI}  & \textbf{F}    & \textbf{P}    & \textbf{R}    & \textbf{\begin{tabular}[c]{@{}c@{}}$N\_clusters$\end{tabular}} \\ 
\hline \hline
0                     & 72.9          & 60.5          & 79.7          & 48.7          & 22                                                                 \\
1                     & 73.1          & 60.7          & 79.7          & 49.0          & 18                                                                \\
2                     & 73.5          & 61.7          & 79.7          & 50.3          & 17                                                                \\
5                     & \textbf{76.7} & \textbf{69.4} & \textbf{79.8} & \textbf{61.3} & 12                                                               \\ \hline
\end{tabular}%
}
\end{table}

\subsubsection{Conclusions on AAS}
As already discussed, for real applications with sparse graphs, the complexity of AAS could be much lower. For all our further experiments, we apply the t-SNE approach \cite{van2014accelerating} to the representations before performing AAS. This is not only computationally simpler and lets us scale to large datasets, but also provides us a good visualization of how the high-dimensional data is clustered. It should be noted that the tSNE embeddings are used only for performing the clustering. The metric is then learned on the original data. It took 1.67 seconds to compute the affinity matrix for the embeddings of MNIST dataset, which consists of $10^4$ examples. For the actual clustering, it took merely $22.823$ seconds on an Intel Core i7-8700 CPU (@ 3.20GHz x 12) with 32 GB RAM. However, due to the large number of hyperparameters, it is important to clearly identify their impact. We make the following conclusions regarding that:
\begin{enumerate}[topsep=-1ex]
    \item In a real application, it is impossible to tune so many hyperparameters in the unsupervised setting. If we have a related dataset, or a separate validation set whose ground truths are available, one may perform clustering using AAS and compute the performance metrics to determine the hyperparameters.
    
    However, our claim in this work is to learn a metric in a completely \textit{unsupervised} manner. Hence, we do not \textit{tune} the hyperparameters for the rest of our experiments (as such, our approach \textit{could} lead to better performance, if further tuned adequately). We merely take guidance from our observations on the MNIST dataset to fix the hyperparameters of AAS apriori.
    \item $k\_graph$, a crucial hyperparameter, when set to $500$ led to a better performance on the MNIST dataset, where each class has $1000$ examples. This indicates that essentially $k\_graph$ should have a value comparable to (around $50\%$) that of the average number of examples per class.
    
    But, for real-world datasets with high overlap (\eg, CUB-200 \cite{CUB}, Cars-196 \cite{Cars196}, JHMDB \cite{JHMDB} etc) where we have fewer examples per class, it is advisable to set $k\_graph$ to a relatively smaller value. For the rest of our experiments, we follow the MOM \cite{MOM} approach, which is also a graph-based approach, and set $k\_graph=50$.
    \item For the other hyperparameters, combinations of $\gamma \in \{10^2,10^3\}$ along with either $(\epsilon \approx 0.65, \theta_{min}=0)$ or $(\epsilon \rightarrow 0.9, \theta_{min} \in \{1,5\} )$ works reasonably well. Having fixed the hyperparameters of AAS, our approach becomes fairly simple to learn a metric, as now it leaves us with only the hyperparameter $\alpha$ in (\ref{eqn_z_i}).
\end{enumerate}

\subsection{Comparison of OPML against state-of-the-art traditional DML methods}
In this subsection, we compare the standalone version of our OPML method against the following state-of-the-art traditional DML methods:
\begin{itemize}
    \item \textbf{JDRML} \cite{JDRML-ICML17}: This method performs Joint-Dimensionality Reduction and Metric Learning (JDRML) to learn a better metric.
    \item \textbf{LRGMML} \cite{LR-GMML}: This method provides a Low-Rank (LR) extension to the state-of-the-art Geometric Mean Metric Learning (GMML) method that learns a metric as a point on the geodesic between two scatter matrices.
    \item \textbf{AML} \cite{AML}: The Adversarial Metric Learning (AML) approach generates synthetic adversarial pairs to provide difficult constraints for learning a metric.
    \item \textbf{MDMLCLDD} \cite{MDMLCLDD}: The Mahalanobis DML with Convex Log-Determinant Divergence (MDMLCLDD) approach provides a convex relaxation to impose orthogonality on the parameters of the distance metric.
\end{itemize}
For all the above methods, we learn a metric using the class labels, as all of them are \textit{supervised} in nature. Using the learned metric we project the test data to the embedding space, where we perform clustering and retrieval on the test embeddings. However, for our method we do not make use of class labels during training. For the comparisons, we make use of the JHMDB, HMDB, AwA2 and Fashion-MNIST datasets, as discussed earlier. We now provide the experimental protocol.

\begin{table}[!htb]
\centering
\caption{Comparison against state-of-the-art traditional supervised DML methods. Bold denotes the best value for a metric.}
\label{SOTA_trad_sup_DML}
\resizebox{0.8\columnwidth}{!}{%
\begin{tabular}{|c|c|cc|cccc|}
\hline
\multicolumn{2}{|c|}{Dataset}                                          & \multicolumn{6}{c|}{\textbf{JHMDB}}                                                           \\ \hline
Method        & \begin{tabular}[c]{@{}c@{}}\textbf{Labels}\end{tabular} & \textbf{NMI}  & \textbf{F}    & \textbf{R@1}  & \textbf{R@2}  & \textbf{R@4}  & \textbf{R@8}  \\ \hline
JDRML         & Yes                                                    & 70.7          & 63.3          & 84.7          & 90.9          & \textbf{95.3} & 97.5          \\
LRGMML        & Yes                                                    & 69.1          & 61.3          & 86.2          & 90.5          & 94.3          & 97.5          \\
AML           & Yes                                                    & 69.5          & 61.4          & \textbf{87.1} & \textbf{91.2} & 94.4          & 97.0          \\
MDML      & Yes                                                    & 69.7          & 62.3          & 87.0          & 91.1          & 94.6          & 97.3          \\ \hline
\textbf{Ours} & No                                                     & \textbf{73.4} & \textbf{67.6} & 86.6          & 89.8          & 94.4          & \textbf{97.8} \\ \hline
\multicolumn{2}{|c|}{Dataset}                                          & \multicolumn{6}{c|}{\textbf{HMDB}}                                                            \\ \hline
Method        & \begin{tabular}[c]{@{}c@{}}\textbf{Labels}\end{tabular} & \textbf{NMI}  & \textbf{F}    & \textbf{R@1}  & \textbf{R@2}  & \textbf{R@4}  & \textbf{R@8}  \\ \hline
JDRML         & Yes                                                    & 48.8          & 28.8          & 72.8          & 81.6          & 88.7          & 93.4          \\
LRGMML        & Yes                                                    & 49.5          & 29.6          & 72.6          & 81.7          & 89.1          & 94.0          \\
AML           & Yes                                                    & \textbf{53.2} & \textbf{31.5} & \textbf{73.3} & \textbf{82.5} & \textbf{89.5} & \textbf{94.8} \\
MDML      & Yes                                                    & 51.2          & 30.3          & 73.2          & 82.4          & 89.4          & 94.5          \\ \hline
\textbf{Ours} & No                                                     & 52.2          & 31.1          & 72.9          & 81.9          & 89.2          & 94.5          \\ \hline
\multicolumn{2}{|c|}{Dataset}                                          & \multicolumn{6}{c|}{\textbf{AwA2}}                                                            \\ \hline
Method        & \begin{tabular}[c]{@{}c@{}}\textbf{Labels}\end{tabular} & \textbf{NMI}  & \textbf{F}    & \textbf{R@1}  & \textbf{R@2}  & \textbf{R@4}  & \textbf{R@8}  \\ \hline
JDRML         & Yes                                                    & 83.2          & 78.7          & 93.9          & 96.9          & 98.8          & 99.4          \\
LRGMML        & Yes                                                    & 82.3          & 77.6          & 94.8          & 97.5          & 98.9          & 99.3          \\
AML           & Yes                                                    & 83.3          & 78.1          & 94.7          & 97.6          & \textbf{99.1} & 99.5          \\
MDML      & Yes                                                    & \textbf{84.3} & \textbf{78.8} & 94.6          & \textbf{97.7} & \textbf{99.1} & 99.6          \\ \hline
\textbf{Ours} & No                                                     & 83.4          & \textbf{78.8} & \textbf{94.9} & 97.6          & 99.0          & \textbf{99.7} \\ \hline
\multicolumn{2}{|c|}{Dataset}                                          & \multicolumn{6}{c|}{\textbf{Fashion-MNIST}}                                                   \\ \hline
Method        & \begin{tabular}[c]{@{}c@{}}\textbf{Labels}\end{tabular} & \textbf{NMI}  & \textbf{F}    & \textbf{R@1}  & \textbf{R@2}  & \textbf{R@4}  & \textbf{R@8}  \\ \hline
JDRML         & Yes                                                    & 59.8          & 47.7          & 77.4          & \textbf{87.2} & 91.7          & 95.6          \\
LRGMML        & Yes                                                    & 59.4          & 49.5          & \textbf{78.6} & 86.4          & \textbf{92.2} & \textbf{96.0} \\
AML           & Yes                                                    & \textbf{64.6} & \textbf{52.5} & 77.8          & 86.5          & 91.5          & 95.2          \\
MDML      & Yes                                                    & 63.9          & 50.2          & 77.7          & 86.3          & 91.6          & 95.3          \\ \hline
\textbf{Ours} & No                                                     & 63.9          & 49.7          & 78.0          & 86.5          & 91.6          & 95.7          \\ \hline
\end{tabular}%
}
\end{table}
\begin{figure}[!htb]
\centering
	\begin{subfigure}{.49\columnwidth}
    	\centering
		\includegraphics[trim={0cm 0cm 0cm 0cm},clip,width=0.8\columnwidth]{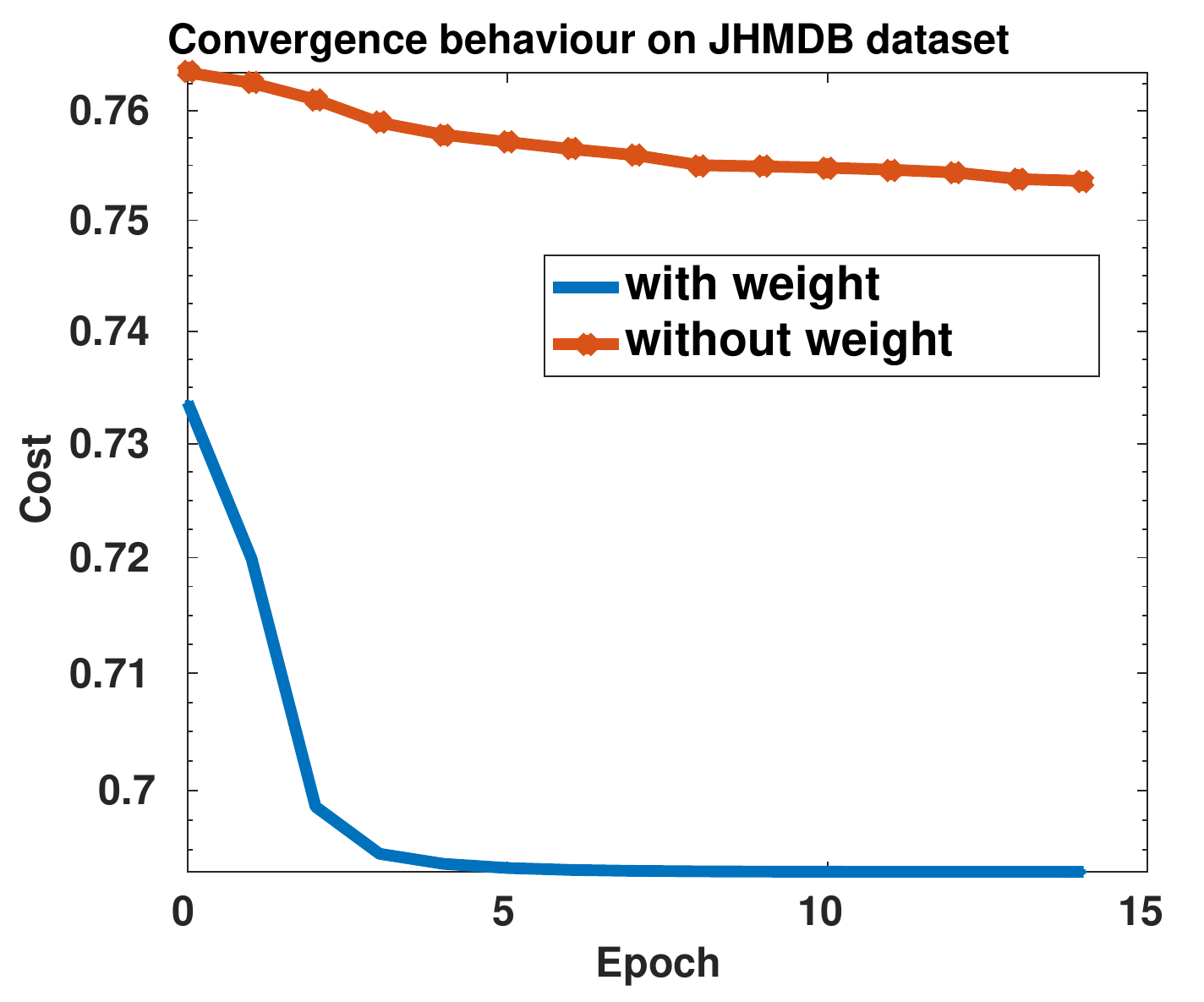}
		\caption{}
        \label{subfig_a}
    \end{subfigure}
	\begin{subfigure}{.49\columnwidth}
    	\centering
		\includegraphics[trim={0cm 0cm 0cm 0cm},clip,width=0.8\columnwidth]{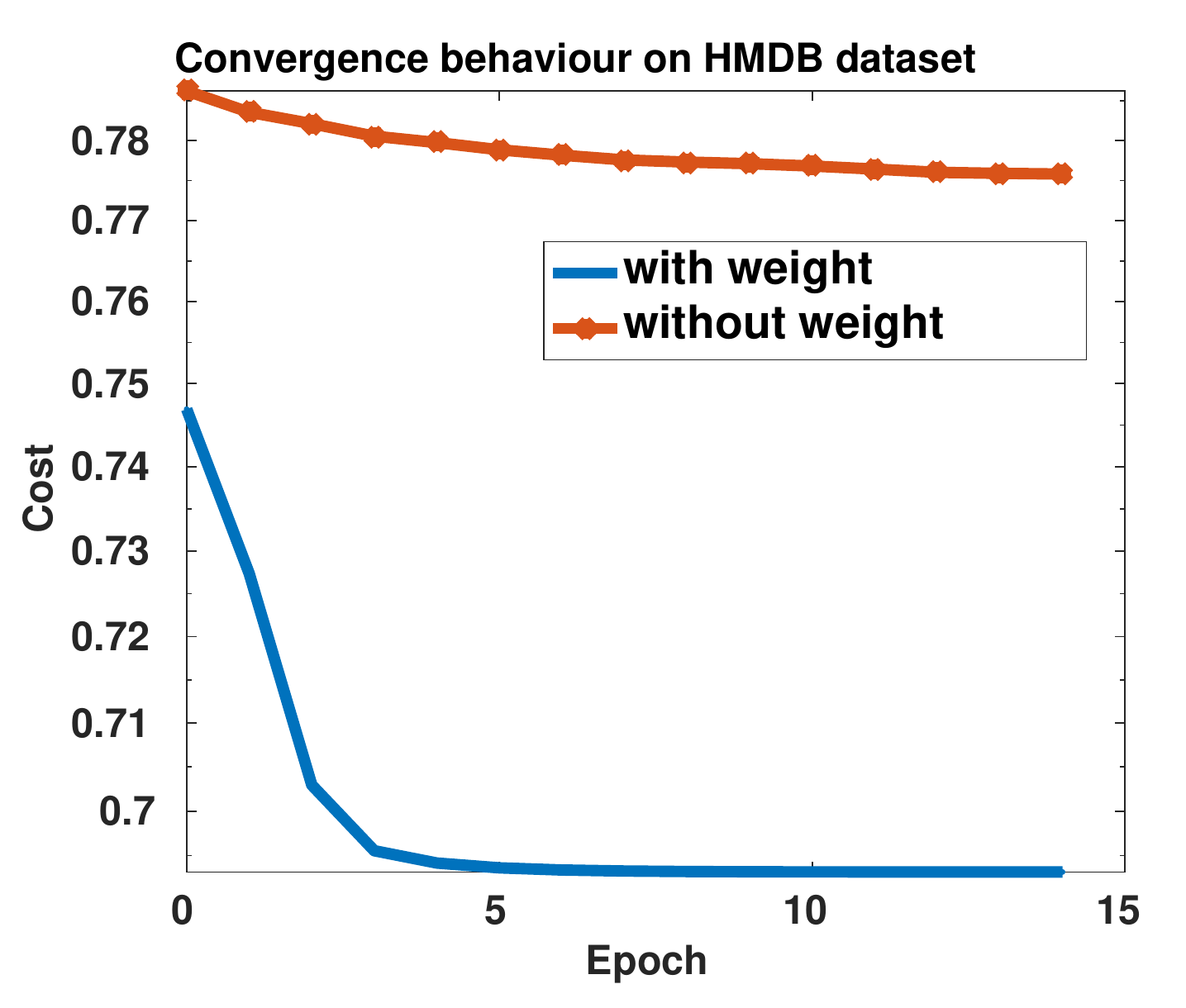}
		\caption{}
        \label{subfig_b}
    \end{subfigure}\\
    \begin{subfigure}{.49\columnwidth}
    	\centering
		\includegraphics[trim={0cm 0cm 0cm 0cm},clip,width=0.8\columnwidth]{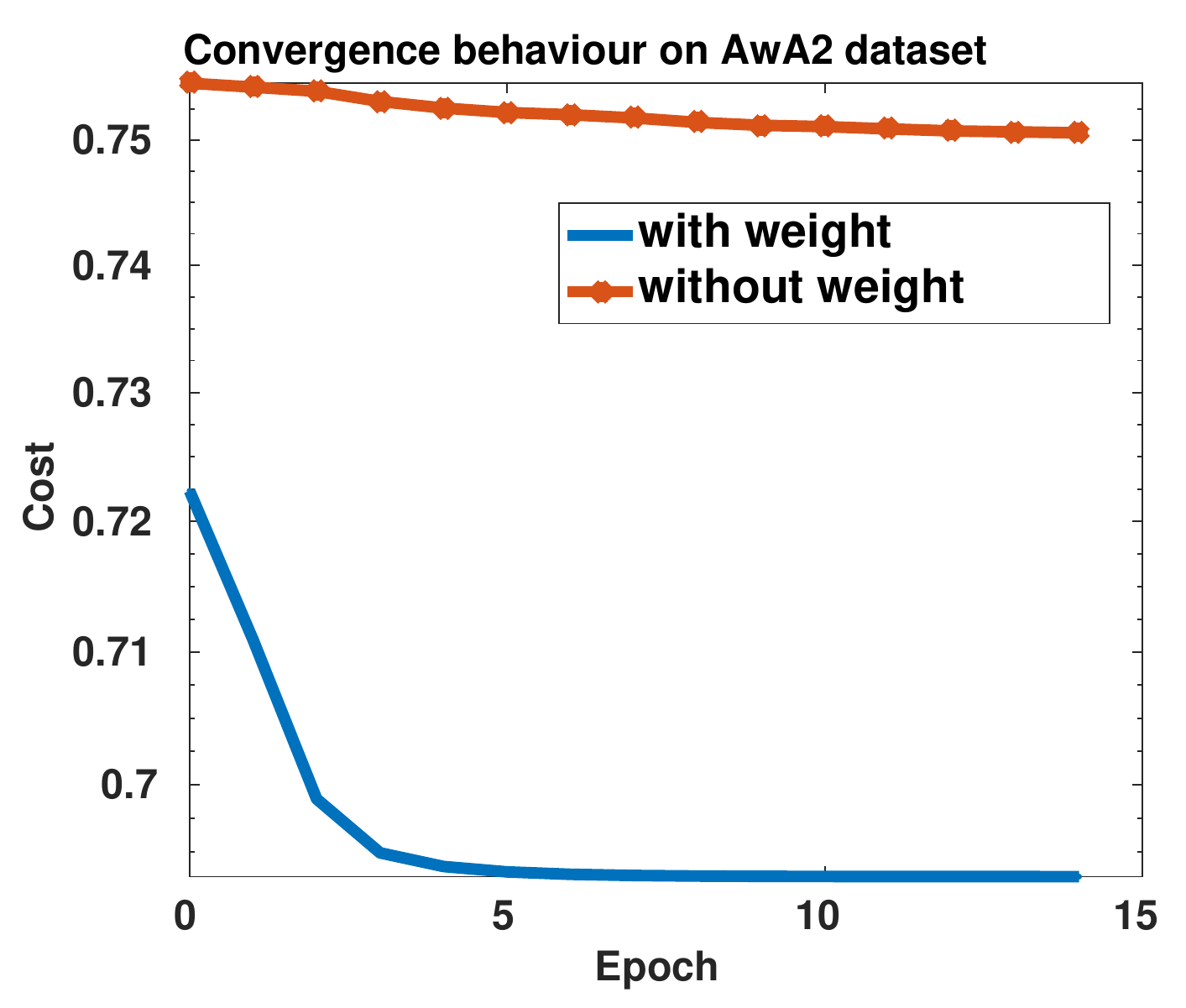}
		\caption{}
        \label{subfig_c}
    \end{subfigure}
    \begin{subfigure}{.49\columnwidth}
    	\centering
		\includegraphics[trim={0cm 0cm 0cm 0cm},clip,width=0.8\columnwidth]{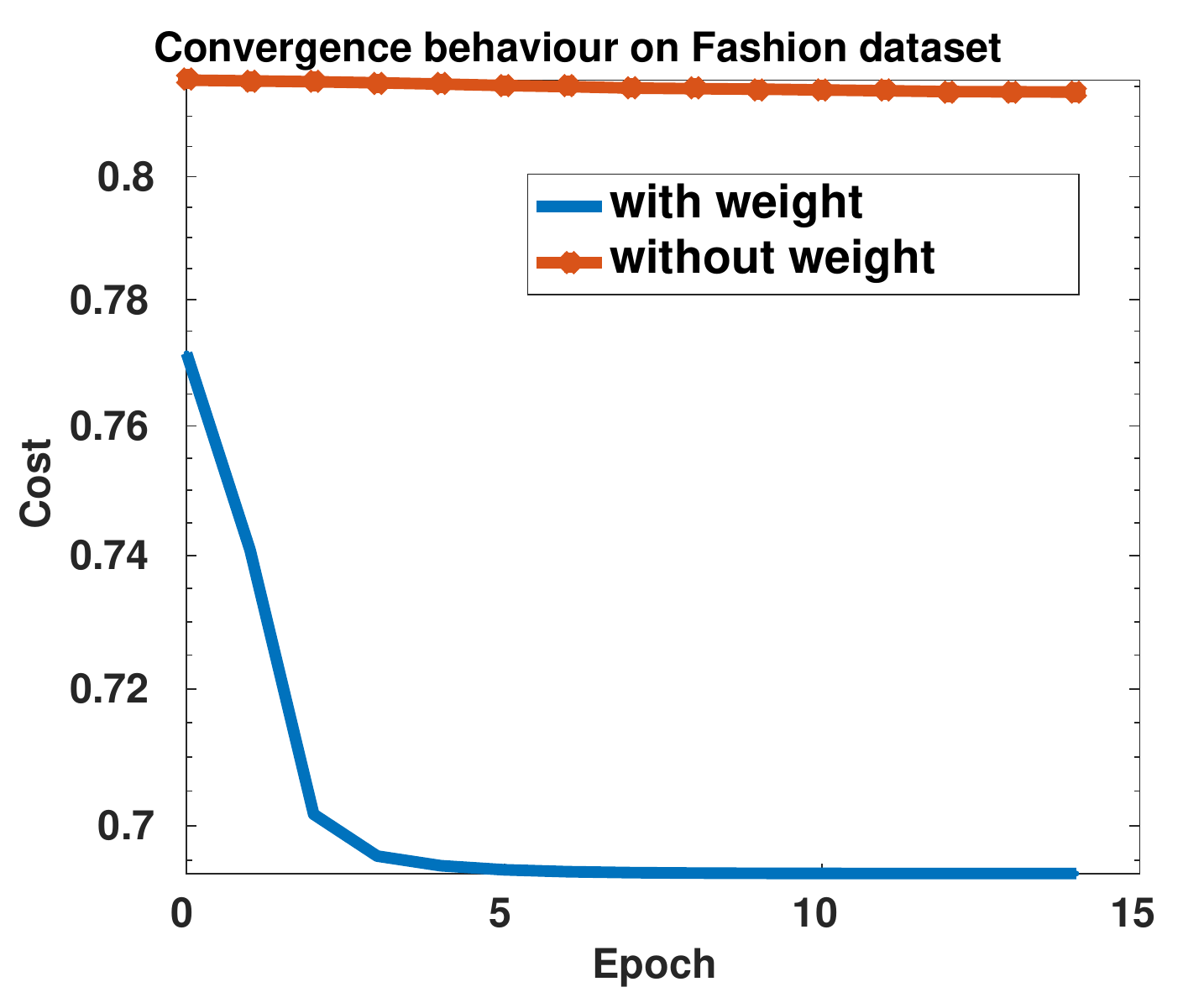}
		\caption{}
        \label{subfig_d}
    \end{subfigure}
    \caption{ Convergence behaviour of our method OPML, and role of the weight function, as observed on: a) JHMDB, b) HMDB, c) AwA2, and d) Fashion-MNIST datasets respectively.}
    \label{ablation_wt}
\end{figure}

\begin{table}[!htb]
\centering
\caption{Sensitivity of OPML towards $\alpha$ on the JHMDB and HMDB datasets, with respect to NMI and F values on the test data.}
\label{ablation_alpha}
\resizebox{0.7\linewidth}{!}{%
\begin{tabular}{|c|c|cccccc|}
\hline
Dataset & $\alpha$ & $35^\circ$ & $40^\circ$ & $45^\circ$ & $50^\circ$ & $55^\circ$ & $60^\circ$ \\ \hline
JHMDB   & NMI      & 69.5       & 70.5       & 73.4       & 71.7       & 72.4       & 68.7       \\
        & F        & 61.4       & 62.2       & 67.6       & 64.8       & 65.3       & 59.0       \\ \hline
HMDB    & NMI      & 51.2       & 52.3       & 52.2       & 52.5       & 52.0       & 51.6       \\
        & F        & 30.8       & 30.9       & 31.1       & 31.0       & 31.0       & 30.4       \\ \hline
\end{tabular}%
}
\end{table}

\subsubsection{Experimental protocol}
For the JHMDB dataset, each frame of a video in this dataset is represented by: i) a RGB vector, and ii) a flow vector. These vectors were provided as part of the work done in Cherian \etal \cite{cherian2018non}. We take average of the \textit{RGB vectors} for all the frames, and represent it as a mean \textit{RGB vector}. We obtain another mean vector obtained by averaging the \textit{flow vectors} for each of the frames. The \textit{mean RGB} and \textit{mean flow} vectors are added together to obtain the final encoding of a video. For the JHMDB dataset, classes 1-8 are for training and 14-21 are for testing.

For the HMDB dataset, we used the split-1 provided as part of the work done in Cherian \etal \cite{cherian2018non}. Similar to JHMDB, we obtain two mean vectors using the representations of all the frames, one for RGB, and the other for flow. These two mean vectors are added to encode a video. For the HMDB dataset, classes 1-21 are for training and 32-51 are for testing. Due to the disjoint nature of training and testing classes in both the datasets, it corresponds to the \textit{zero-shot learning} scenario.

We now design experiments on the AwA2 and Fashion-MNIST datasets to highlight the robustness to noise. For the AwA2 dataset, the features provided in Xian \etal \cite{ZSL_good_bad_ugly} have been used in our experiments. The ten largest classes have been picked for our experiments. Based on descending order of the number of examples present in the selected classes, following are the class ids: 7, 23, 40, 49, 31, 27, 38, 29, 1, and 19. From each class, examples 1-200 are used for training, and examples 401-600 are used for testing. To tune the compared supervised baselines, examples 201-400 are used as validation data. To act as noise for the AwA2 dataset, images from the first 10 classes of the CUB dataset \cite{CUB} have been added. The features for CUB are the same as in AwA2, and are provided by Xian \etal \cite{ZSL_good_bad_ugly}.

For the Fashion-MNIST dataset, the first 200 examples from each class of the standard training split has been chosen to constitute our training data. To tune the compared supervised baselines, the next 200 examples from each class are used as validation data. The first 200 examples from each class of the standard testing split has been chosen to constitute our test data. To act as noise for the Fashion dataset, we perform the following: i) The first 150 digit images from each class of the original MNIST dataset is picked. ii) To each pixel of the chosen digit images, we add random noise. iii) This collection of noisy digit images has been added along with the Fashion dataset. The raw pixels have been concatenated to form the feature vectors.

In all datasets, the examples have been $l2$-normalized.

\subsubsection{Optimization}
We performed Riemannian Conjugate Gradient Descent (RCGD) using the Manopt toolbox \cite{manopt}. We kept all the default Manopt parameters (\eg, learning rate), except the maximum number of epochs of RCGD, which we set as 30. Due to the product manifold based optimization, our method converges quite fast (Figure \ref{ablation_wt}). This is in accordance with studies in Riemannian optimization \cite{RiemSVRG_NIPS16,bonnabel2013stochastic}.

\subsubsection{Empirical Observations}
In Table \ref{SOTA_trad_sup_DML}, we report the results of the compared approaches. For our method, we arbitrarily set $\alpha=45^\circ$, and embedding size of 128. We also indicate the nature of an approach, \ie, whether it makes use of class labels or not. Our method performs competitive, despite not making use of class labels. This shows promise of the probabilistic objective (\ref{OPML}) of our method to provide a novel way to learn a metric, while being unsupervised in nature.

\begin{table*}[t]
\centering
\caption{Comparisons on FGVC datasets. The best value for each metric is shown in bold, while the second best is shown in underline.}
\label{FGVC_all}
\resizebox{0.9\linewidth}{!}{%
\begin{tabular}{|c|c|ccccc|ccccc|cccc|}
\hline
\multicolumn{2}{|c|}{\textbf{Dataset}}       & \multicolumn{5}{c|}{\textbf{CUB 200} \cite{CUB}}                                         & \multicolumn{5}{c|}{\textbf{Cars 196} \cite{Cars196}}                                        & \multicolumn{4}{c|}{\textbf{SOP} \cite{Lifted_structure}}         \\ \hline
\textbf{Method}      & \textbf{Labels} & \textbf{NMI}  & \textbf{R@1}  & \textbf{R@2}  & \textbf{R@4}  & \textbf{R@8}  & \textbf{NMI}  & \textbf{R@1}  & \textbf{R@2}  & \textbf{R@4}  & \textbf{R@8}  & \textbf{NMI}  & \textbf{R@1}  & \textbf{R@10} & \textbf{R@100} \\ \hline
Initial (Random)    & No                    & 34.6          & 31.5          & 42.4          & 55.0          & 67.0          & 23.4          & 22.7          & 31.7          & 42.7          & 54.8          & 79.8          & 25.4          & 35.6          & 48.5           \\ \hline
DeepCluster \cite{DeepCluster_ECCV18}         & No                    & 53.0          & 42.9          & 54.1          & 65.6          & 76.2          & 38.5          & 32.6          & 43.8          & 57.0          & 69.5          & 82.8          & 34.6          & 52.6          & 66.8           \\
MOM \cite{MOM}                 & No                    & 55.0 & 45.3 & 57.8 & 68.6 & 78.4          & \textbf{38.6}          & 35.5          & 48.2          & 60.6          & 72.4          & 84.4          & 43.3          & 57.2          & 73.2           \\
Exemplar \cite{ExemplarCNN_TPAMI16}             & No                    & 45.0          & 38.2          & 50.3          & 62.8          & 75.0          & 35.4          & 36.5          & 48.1          & 59.2          & 71.0          & \underline{85.0}          & 45.0          & 60.3          & 75.2           \\
InvariantSpread \cite{InvariantSpread_CVPR19}                 & No                    & \underline{55.4} & \underline{46.2} & \underline{59.0} & \underline{70.1} & \underline{80.2}          & 35.8          & \underline{41.3}          & \underline{52.3}          & \underline{63.6}          & \underline{74.9}          & \textbf{86.0}          & \textbf{48.9}          & \textbf{64.0}          & \textbf{78.0}           \\\hline
\textbf{Ours} & No                    & \textbf{55.6}          & \textbf{47.1}          & \textbf{59.7}          & \textbf{72.1}          & \textbf{82.8} & \underline{38.1} & \textbf{45.0} & \textbf{56.2} & \textbf{66.7} & \textbf{76.6} & 84.2          & \underline{45.5} & \underline{61.6} & \underline{77.1}  \\ \hline
\end{tabular}%
}
\end{table*}

\subsection{Ablation studies studying the role of the pseudo-label mining approach, weight function, and sensitivity to hyperparameter $\alpha$ in the probabilistic objective}
Having demonstrated the promise of our proposed orthogonality-based probabilistic objective, we conducted a few experiments to observe the effect of replacing the AAS method with k-means for obtaining pseudo-labels, while setting $k$ to the actual number of classes. By replacing AAS with k-means on the HMDB dataset, we observed a value drop of 6.8 for NMI, 1.2 in R@1, 0.4 in R@2, and 0.5 in R@4 and R@8. Similarly we observed a value drop of 2.4 in NMI, 0.5 in R@1 and 1.6 in R@8, by replacing AAS with k-means on the JHMDB dataset. Interestingly, our method gained a value of 0.5 in NMI by replacing AAS with k-means on MNIST-Fashion, as well as gain of 2.2 in NMI and 0.4 in R@1 upon replacing AAS with k-means on the AwA2 dataset.

This implies that our probabilistic objective could still be used in conjunction with any alternative pseudo-label mining approach. However, exploration of such alternatives is beyond the scope of this paper, as our focus is on the metric learning objective. In our paper, we pick the AAS clustering approach for its robustness, intuitive nature, and conceptual superiority over the naive k-means approach, as shown earlier for certain scenarios. The major drawback of the k-means is its dependency on prior knowledge about the number of classes in data, which is infeasible in the purely unsupervised setting.

We also report the convergence behaviour of our method OPML, both with and without the weight function, in Figure \ref{ablation_wt}. The weight function leads to lower values of the objective and a faster convergence. In Table \ref{ablation_alpha}, we report the performance of our method on the JHMDB and HMDB datasets, against varying values of $\alpha$ as present in (\ref{eqn_z_i}). We observed that the performances are fairly stable, in the range $35^\circ-60^\circ$.

\subsection{Studies on Fine-Grained Visual Categorization (FGVC)}

We now compare the proposed stochastic version of our method (sOPML) against the following unsupervised deep approaches for feature learning: DeepCluster \cite{DeepCluster_ECCV18}, MOM \cite{MOM}, Exemplar \cite{ExemplarCNN_TPAMI16} and InvariantSpread \cite{InvariantSpread_CVPR19}. As discussed in Section \ref{sec_rel_work}, these approaches are representative of the various paradigms of unsupervised metric learning. DeepCluster makes use of pseudo-labels using k-means clustering, MOM is a graph-based method utilizing random walk to identify triplets, Exemplar and InvariantSpread are instance-wise methods. We compare the approaches on the CUB-200, Cars-196 and SOP datasets as discussed earlier. These datasets are fine-grained in nature with huge inter-class similarities and intra-class variances, and are standard benchmarks in evaluating deep metric learning approaches, due to the challenging nature of images present in them. The train-test splits mentioned earlier follows standard deep metric learning protocol \cite{Lifted_structure}.

We used GoogLeNet \cite{GoogLeNet} pretrained on ImageNet \cite{ImageNet2015}, as the backbone CNN, using the MatConvNet \cite{MatConvNet} tool, and follow the standard protocol for training (on a Tesla V100-PCIE-16GB). Following the graph-based label mining approach MOM \cite{MOM}, the initial features for AAS are formed by the Regional Maximum Activation of Convolutions (R-MAC) \cite{RMAC_ICLR16} right before the average pool layer, and aggregated over three input scales ($512$, $512/\sqrt{2}$, $256$). Similar to the training data, we also used the same scales to obtain the final R-MAC embeddings of the test examples using the learned model. We used mini-batch size of 120 and set $maxiter=10$ in Algorithm \ref{alg_sOPML}, \ie, 10 RCGD updates are done for each mini-batch in the loss layer. All other parameters are kept the same as default. The embedding size has been kept as 128, and we set $\alpha=45^\circ$. The test embeddings are used for evaluating the clustering and retrieval performances of the compared approaches.

\begin{figure}[!htb]
\vspace{-0.5cm}
\begin{subfigure}{\columnwidth}
    	\centering
		\includegraphics[width=0.7\columnwidth]{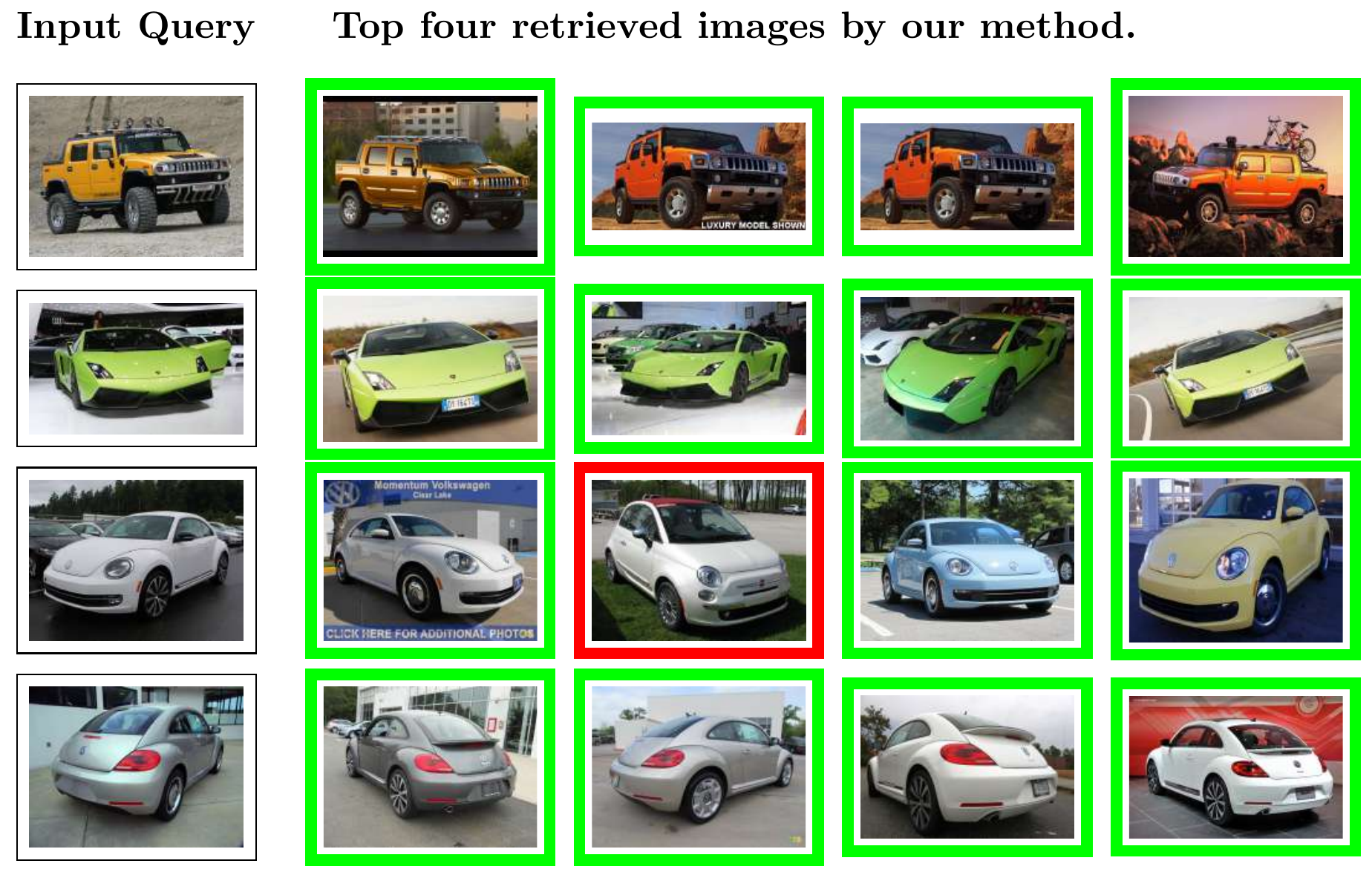}
		\caption{Successful cases.}
         \label{retrieve_success}
\end{subfigure}\\
\begin{subfigure}{\columnwidth}
    	\centering
		\includegraphics[width=0.7\columnwidth]{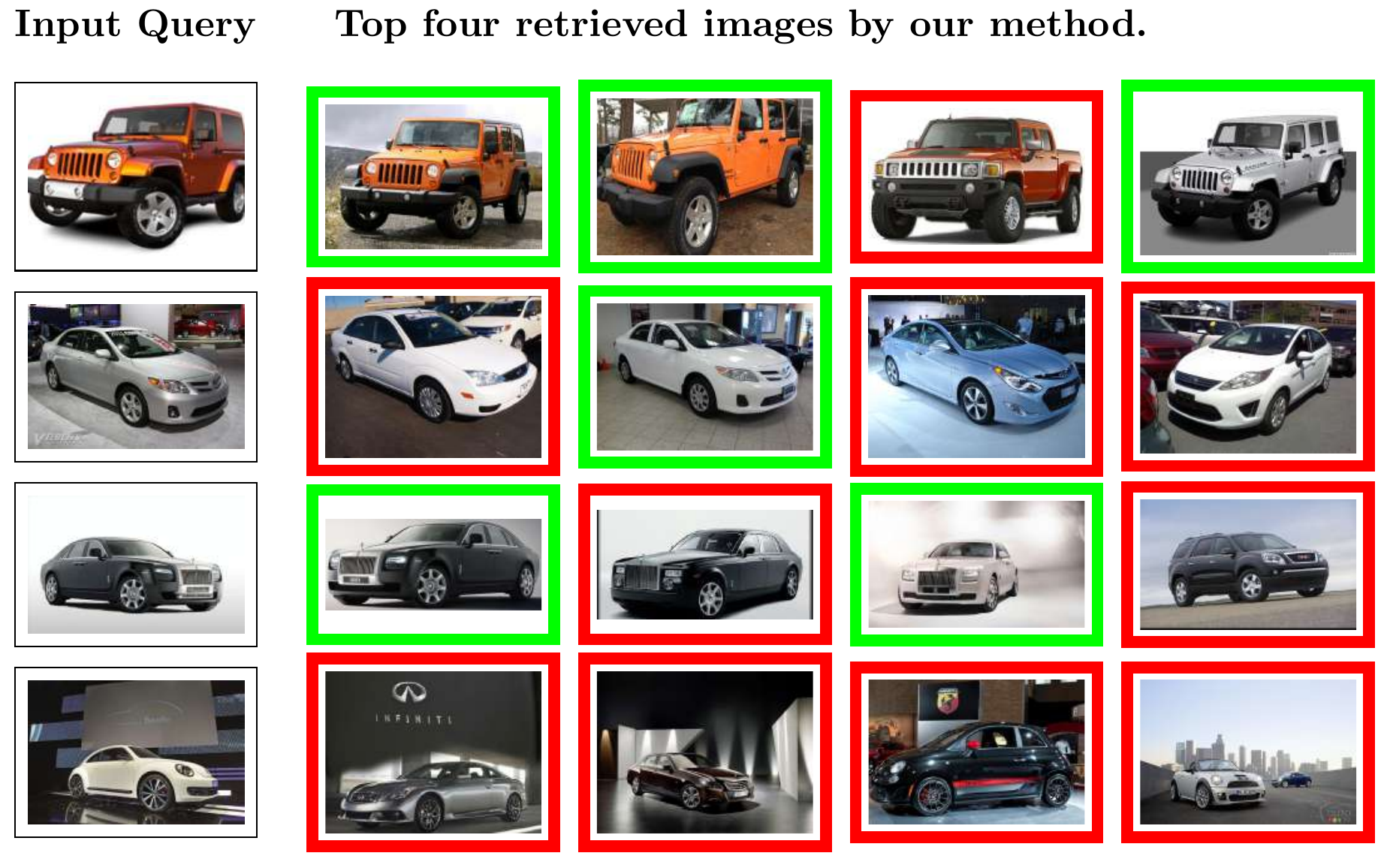}
		\caption{Failure cases.}
         \label{retrieve_failure}
\end{subfigure}
\caption{(a-b) Retrieval results of our method on the Cars-196 dataset (Best viewed in color). For a query image, a retrieved image is shown in green if it is correct (from same class), and in red if it is otherwise.}
\vspace{-0.6cm}
\end{figure}

The results on the FGVC datasets are reported in Table \ref{FGVC_all}. Our method achieves state-of-the-art performance on the CUB-200 and Cars-196 datasets. On the SOP dataset, where each class has extremely few examples, the InvariantSpread approach by virtue of its augmentation and spread-out properties performs the best. In Figure \ref{retrieve_success}-\ref{retrieve_failure}, we present a few retrieval results of our method on the Cars-196 dataset. Additionally, Table \ref{vs_sup_deepDML} compares our method against a few supervised deep baselines.

\begin{table*}[t]
\centering
\caption{Ablation study comparing different variants of our method against the MOM approach.}
\label{FGVC_vs_MOM}
\resizebox{0.85\linewidth}{!}{%
\begin{tabular}{|c|c|ccccc|ccccc|cccc|}
\hline
\multicolumn{2}{|c|}{\textbf{Dataset}}       & \multicolumn{5}{c|}{\textbf{CUB 200} \cite{CUB}}                                         & \multicolumn{5}{c|}{\textbf{Cars 196} \cite{Cars196}}                                        & \multicolumn{4}{c|}{\textbf{SOP} \cite{Lifted_structure}}         \\ \hline
\textbf{Method}      & \textbf{Labels} & \textbf{NMI}  & \textbf{R@1}  & \textbf{R@2}  & \textbf{R@4}  & \textbf{R@8}  & \textbf{NMI}  & \textbf{R@1}  & \textbf{R@2}  & \textbf{R@4}  & \textbf{R@8}  & \textbf{NMI}  & \textbf{R@1}  & \textbf{R@10} & \textbf{R@100} \\ \hline
MOM \cite{MOM}                 & No                    & 55.0 & 45.3 & 57.8 & 68.6 & 78.4          & \textbf{38.6}          & 35.5          & 48.2          & \underline{60.6}          & \underline{72.4}          & \textbf{84.4}          & 43.3          & 57.2          & 73.2           \\
Ours-wo\_ortho              & No                    &\underline{55.5}	&\underline{46.7}	&\underline{59.2}	&\underline{71.7}	&\underline{82.6} &36.8	&\underline{38.8}	&\underline{49.4}	&60.1	&71.1 &82.5	&\underline{43.8}	&\underline{59.4}	&\underline{74.5}          \\ \hline
\textbf{Ours} & No                    & \textbf{55.6}          & \textbf{47.1}          & \textbf{59.7}          & \textbf{72.1}          & \textbf{82.8} & \underline{38.1} & \textbf{45.0} & \textbf{56.2} & \textbf{66.7} & \textbf{76.6} & \underline{84.2}          & \textbf{45.5} & \textbf{61.6} & \textbf{77.1}  \\ \hline
\end{tabular}%
}
\end{table*}
\begin{table}[!htb]
\centering
\caption{Comparison against the graph-based MOM method, and ablation studies with respect to orthogonality.}
\label{MOM_ablation_ortho}
\resizebox{0.8\columnwidth}{!}{%
\begin{tabular}{|c|c|cc|cccc|}
\hline
\multicolumn{2}{|c|}{Dataset}                                           & \multicolumn{6}{c|}{\textbf{JHMDB}}                                                           \\ \hline
Method         & \begin{tabular}[c]{@{}c@{}}\textbf{Labels}\end{tabular} & \textbf{NMI}  & \textbf{F}    & \textbf{R@1}  & \textbf{R@2}  & \textbf{R@4}  & \textbf{R@8}  \\ \hline
MOM            & No                                                     & 65.2          & 57.1          & 80.9          & 88.2          & \textbf{94.5} & 96.5          \\
Ours-wo\_ortho & No                                                     & 66.6          & 53.6          & 85.2          & \textbf{90.3} & 94.1          & 96.2          \\ \hline
\textbf{Ours}  & No                                                     & \textbf{73.4} & \textbf{67.6} & \textbf{86.6} & 89.8          & 94.4          & \textbf{97.8} \\ \hline
\multicolumn{2}{|c|}{Dataset}                                           & \multicolumn{6}{c|}{\textbf{HMDB}}                                                            \\ \hline
Method         & \begin{tabular}[c]{@{}c@{}}\textbf{Labels}\end{tabular} & \textbf{NMI}  & \textbf{F}    & \textbf{R@1}  & \textbf{R@2}  & \textbf{R@4}  & \textbf{R@8}  \\ \hline
MOM            & No                                                     & 41.4          & 24.4          & 67.1          & 75.9          & 84.2          & 91.7          \\
Ours-wo\_ortho & No                                                     & 47.8          & 28.7          & 70.5          & 80.5          & 87.9          & 93.8          \\ \hline
\textbf{Ours}  & No                                                     & \textbf{52.2} & \textbf{31.1} & \textbf{72.9} & \textbf{81.9} & \textbf{89.2} & \textbf{94.5} \\ \hline
\multicolumn{2}{|c|}{Dataset}                                           & \multicolumn{6}{c|}{\textbf{AwA2}}                                                            \\ \hline
Method         & \begin{tabular}[c]{@{}c@{}}\textbf{Labels}\end{tabular} & \textbf{NMI}  & \textbf{F}    & \textbf{R@1}  & \textbf{R@2}  & \textbf{R@4}  & \textbf{R@8}  \\ \hline
MOM            & No                                                     & 68.1          & 62.2          & 87.1          & 93.3          & 96.6          & 98.7          \\
Ours-wo\_ortho & No                                                     & 73.6          & 65.4          & 93.5          & 97.5          & 98.7          & 99.1          \\ \hline
\textbf{Ours}  & No                                                     & \textbf{83.4} & \textbf{78.8} & \textbf{94.9} & \textbf{97.6} & \textbf{99.0} & \textbf{99.7} \\ \hline
\multicolumn{2}{|c|}{Dataset}                                           & \multicolumn{6}{c|}{\textbf{Fashion-MNIST}}                                                   \\ \hline
Method         & \begin{tabular}[c]{@{}c@{}}\textbf{Labels}\end{tabular} & \textbf{NMI}  & \textbf{F}    & \textbf{R@1}  & \textbf{R@2}  & \textbf{R@4}  & \textbf{R@8}  \\ \hline
MOM            & No                                                     & 53.7          & 42.2          & 73.1          & 82.3          & 89.3          & 94.8          \\
Ours-wo\_ortho & No                                                     & 54.8          & 42.5          & 76.6          & 84.6          & \textbf{91.7} & 95.3          \\ \hline
\textbf{Ours}  & No                                                     & \textbf{63.9} & \textbf{49.7} & \textbf{78.0} & \textbf{86.5} & 91.6          & \textbf{95.7} \\ \hline
\end{tabular}%
}
\end{table}

\begin{table}[!htb]
\centering
\caption{Comparison against supervised deep metric learning methods on the FGVC datasets.}
\label{vs_sup_deepDML}
\resizebox{0.75\columnwidth}{!}{%
\begin{tabular}{|c|c|c|cccc|}
\hline
\textbf{Dataset}                   & \textbf{Method} & \textbf{Labels} & \textbf{R@1} & \textbf{R@2} & \textbf{R@4}  & \textbf{R@8}   \\ \hline
\multirow{4}{*}{\textbf{CUB 200}}  & Triplet-SH \cite{FaceNet}         & Yes             & 40.6         & 52.3         & 64.2          & 75.0           \\
                          & Angular \cite{Angular_loss}        & Yes             & 53.6         & 65.0         & 75.3          & 83.7           \\
                          & Multi-Sim \cite{multisim_CVPR19}      & Yes             & 65.7         & 77.0         & 86.3          & 91.2           \\ \cline{2-7} 
                          & Ours            & No              & 47.1         & 59.7         & 72.1          & 82.8           \\ \hline
\textbf{Dataset}                   & \textbf{Method} & \textbf{Labels} & \textbf{R@1} & \textbf{R@2} & \textbf{R@4}  & \textbf{R@8}   \\ \hline
\multirow{4}{*}{\textbf{Cars 196}} & Triplet-SH \cite{FaceNet}         & Yes             & 53.2         & 65.4         & 74.3          & 83.6           \\
                          & Angular \cite{Angular_loss}         & Yes             & 71.3         & 80.7         & 87.0          & 91.8           \\
                          & Multi-Sim \cite{multisim_CVPR19}       & Yes             & 84.1         & 90.4         & 94.0          & 96.5           \\ \cline{2-7} 
                          & Ours            & No              & 45.0         & 56.2         & 66.7          & 76.6           \\ \hline
\textbf{Dataset}                   & \textbf{Method} & \multicolumn{2}{c|}{\textbf{Labels}}    & \textbf{R@1} & \textbf{R@10} & \textbf{R@100} \\ \hline
\multirow{4}{*}{\textbf{SOP}} & Triplet-SH \cite{FaceNet}         & \multicolumn{2}{c|}{Yes}              & 57.8         & 75.3          & 88.1           \\
                          & Angular \cite{Angular_loss}         & \multicolumn{2}{c|}{Yes}              & 67.9         & 83.2          & 92.2           \\
                          & Multi-Sim \cite{multisim_CVPR19}       & \multicolumn{2}{c|}{Yes}              & 78.2         & 90.5          & 96.0           \\ \cline{2-7} 
                          & Ours            & \multicolumn{2}{c|}{No}              & 45.5         & 61.6          & 77.1           \\ \hline
\end{tabular}%
}
\end{table}

\subsection{Studies comparing against MOM, with respect to orthogonality, and role of the probabilistic objective}
We performed experiments to study the role of the orthogonality constraint in our approach. A look at Table \ref{MOM_ablation_ortho} demonstrates that the main benefit of orthogonality comes into the picture when considering the clustering performance, as also expected from Figure \ref{ortho_motiv}. In that case, it is essential to avoid a model collapse (leading to many examples having nearby embeddings). For the retrieval task, it suffices to have just one correct example from the same class within the top K retrieved examples to boost the Recall@K value. This implies that one may not observe a profound impact of orthogonality for retrieval. However, including orthogonality does lead to further improvements in general. Table \ref{FGVC_vs_MOM} shows the benefit of orthogonality in the FGVC task. On Cars-196, we observed significant gains in retrieval performance as well.

Using Table \ref{MOM_ablation_ortho}, we would also like to highlight the gains obtained by our method over MOM, just by using our proposed probabilistic loss in (\ref{OPML}), without using orthogonality (Ours-wo\_ortho). When compared against MOM, using our loss alone our method achieves: i) 4.3 better R@1 on JHMDB, ii) 3.4, 4.6, 3.7, 2.1 better R@1, R@2, R@4 and R@8 respectively, on HMDB, iii) 6.4, 4.2, 2.1, 0.4 better R@1, R@2, R@4 and R@8 respectively, on AwA2, and iv) 3.5, 2.3, 2.4, 0.5 better R@1, R@2, R@4 and R@8 respectively, on Fashion-MNIST. Similarly, just by using our loss, we outperform MOM consistently in the FGVC datasets as well (Table \ref{FGVC_vs_MOM}).

\section{Conclusion and Future Work}
In this paper, we propose an approach to learn a discriminative distance metric without using manually annotated class labels. The motivation for our paper is the infeasibility of obtaining class labels in many crucial machine learning applications. In particular, we proposed a novel orthogonality-based probabilistic loss for metric learning, that inherently seeks to preserve an angular constraint on a triplet of examples obtained using pseudo-labels. In our paper, we employ a graph-based clustering method to obtain pseudo-labels due to its intuitive and conceptual prowess. However, the same can be replaced with an alternative pseudo-label mining method. The orthogonality component in our method avoids a model collapse of the embeddings obtained, and in general leads to a better metric. In future, our method could be extended along with a complementary augmentation-based self-supervised approach to obtain the initial representations.

{\small
\bibliographystyle{unsrt}
\bibliography{OPML_TAI_arXiv20}
}

\end{document}